\documentclass[sn-mathphys-num]{sn-jnl}
\usepackage{amsmath,amssymb,amsfonts}
\usepackage{algorithmic}
\usepackage{graphicx}
\usepackage{textcomp}
\usepackage{placeins}
\usepackage{xcolor}
\usepackage{amsmath}
\usepackage{multirow}
\usepackage{booktabs}
\usepackage{verbatim}
\usepackage{hyperref}
\usepackage[table]{xcolor} 
\usepackage{multirow}
\usepackage[british]{babel}

\graphicspath{{./figures/}}
\usepackage[dvipsnames]{xcolor} 

\newcommand{\revd}[1]{\textcolor{black}{#1}}

\newcommand{\revdg}[1]{\textcolor{black}{#1}}
\begin{document}

\title[Loosely coupled 4D-Radar-Inertial Odometry\\ for Ground Robots]{Loosely coupled 4D-Radar-Inertial Odometry\\ for Ground Robots}


\author*[1]{\fnm{Lucía} \sur{Coto-Elena}}\email{lcotele@upo.es}
\equalcont{These authors contributed equally to this work.}

\author[1]{\fnm{Fernando} \sur{Caballero}}\email{fcaballero@upo.es}
\equalcont{These authors contributed equally to this work.}

\author[1]{\fnm{Luis} \sur{Merino}}\email{lmercab@upo.es}
\equalcont{These authors contributed equally to this work.}

\affil*[1]{\orgdiv{Service Robotics Laboratory}, \orgname{Universidad Pablo de Olavide}, \orgaddress{\city{Sevilla}, \postcode{41013}, \country{Spain}}}
\maketitle

{\centering\scriptsize
Published in \textit{Journal of Intelligent and Robotic Systems}, Volume 111, Article 107 (2025). \\
DOI: \href{https://doi.org/10.1007/s10846-025-02301-9}{10.1007/s10846-025-02301-9}.\par
}

\abstract{Accurate robot odometry is essential for autonomous navigation. While numerous techniques have been developed based on various sensor suites, odometry estimation using only radar and IMU remains an underexplored area. Radar proves particularly valuable in environments where traditional sensors, like cameras or LiDAR, may struggle, especially in low-light conditions or when faced with environmental challenges like fog, rain or smoke. However, despite its robustness, radar data is noisier and more prone to outliers, requiring specialized processing approaches. In this paper, we propose a graph-based optimization approach (\url{https://github.com/robotics-upo/4D-Radar-Odom.git})  using a sliding window for radar-based odometry, designed to maintain robust relationships between poses by forming a network of connections, while keeping computational costs fixed (specially beneficial in long trajectories).  Additionally, we introduce an enhancement in the ego-velocity estimation specifically for ground vehicles, both holonomic and non-holonomic, which subsequently improves the direct odometry input required by the optimizer. Finally, we present a comparative study of our approach against existing algorithms, showing how our pure odometry approach inproves the state of art in \revdg{all} trajectories of the NTU4DRadLM dataset, achieving promising results when evaluating key performance metrics.}

\keywords{Ground robots, 4D Radar-Inertial Odometry,  Doppler Velocity.}

\maketitle

\section{Introduction}
\label{Introduction}

In autonomous navigation, the ability of vehicles to self-localize in unknown environments is essential, especially when external information systems like GPS or environmental maps are unavailable or insufficiently precise. Developing algorithms that can estimate the vehicle's trajectory based solely on local sensor information remains an active area of research.

The most accurate methods often involve multi-modal sensor fusion, combining data from sensors like LiDARs and cameras to leverage their strengths. These approaches have shown strong performance by taking advantage of the dense and precise data these sensors provide \cite{legoloam2018,yan2023gs,openvslam2019,min2021voldorslamtimesfeaturebaseddirect}. However, despite their advantages, both LiDARs and cameras present limitations that restrict their use in certain environments. Cameras, while affordable and high-resolution, struggle in low-visibility conditions or poor lighting and face challenges in measuring depth in large and open areas. They are also highly sensitive to environmental factors such as fog, rain, or snow, which can significantly degrade their performance. LiDARs, known for their precise distance measurements, also face challenges in adverse weather conditions, where particles like dust, fog, or rain scatter the laser pulses, generating false or incomplete data \cite{zywanowski2020}.
Radar technology emerges as a promising alternative due to its ability to operate in adverse environmental conditions like fog, dust, or low lighting\cite{burnett2022}. Its longer wavelengths allow it to penetrate these obstacles without being affected, and its capability to measure velocity through the Doppler effect provides valuable information for motion estimation \cite{ITSC2013}. 

Despite its advantages, radar presents several challenges that differentiate it from other sensors like LiDAR, with which it is commonly compared due to the similarity in data format—point clouds\cite{Chan_2023}. Radar data is inherently noisier and sparser,  leading to point clouds with more irrelevant or incorrect information, complicating precise environmental interpretation. Its lower spatial and angular resolution also limits radar’s ability to capture fine structural details, making feature extraction techniques more challenging. Additionally, radar data is affected by artifacts such as specular reflections, multipath effects, and signal attenuation,  leading to false or missed detections. Specular reflections\textbf{ }occur when radar waves bounce off reflective surfaces at specific angles, causing false detections where objects appear in incorrect positions or missed detections when the signal fails to return. Multipath effects arise when signals reflect off multiple surfaces before returning to the radar, creating multiple echoes for the same object and distorting its perceived location and distance, with signal strength decreasing after each reflection. Signal attenuation also impacts radar reliability, as certain materials can absorb radar waves rather than reflecting them, resulting in additional missed detections. These artifacts contribute to flickering between adjacent frames and ghost reflections, resulting in an unstable and potentially misleading perception of the environment.
\cite{9760734,10530463}.

These challenges prevent the direct application of LiDAR-based techniques to radar data. Therefore, specialized algorithms are required to process radar's noisier, lower-resolution data while mitigating artifacts that compromise data stability. \revd{One critical aspect affected by radar noise is ego-velocity estimation, particularly in the vertical component, where lower spatial resolution and multipath effects contribute to increased drift and instability. Accurately estimating velocity from radar measurements is essential for reliable vehicle localization and motion prediction.}

In this paper, we present an odometry system that enables effective vehicle navigation in challenging environments using radar sensor and an Inertial Measurement Unit (IMU). The system mitigates the effects of inherent sensor noise through a novel methodology for estimating the vehicle velocity via the Doppler effect, followed by a robust odometry estimation framework based on Pose Graph optimization. \revd{Therefore, our main contributions are:}

\begin{itemize} 
\item \revd{A methodology for ego-velocity estimation that achieves higher accuracy by reducing the impact of radar noise on velocity estimation, specifically minimizing vertical drift. This method incorporates motion constraints derived from the vehicle’s kinematics (e.g., vertical movement restrictions in ground vehicles) directly into the estimation model. With IMU orientation data integrated between successive samples, the approach enhances the reliability of the velocity estimation. Additionally, RANSAC filtering mitigates the impact of dynamic objects, further refining accuracy.}
\item \revd{An odometry system designed with pose graph optimization that fuses radar and IMU measurements within a sliding window of the most recent \textit{n} keyframes. Within this framework, poses in the window are interconnected, forming a mesh that establishes relative and local relationships. This optimization approach refines the vehicle's trajectory by incorporating constraints from both IMU data and the alignment of radar scans at each keyframe. By leveraging these interconnections, the system enhances robustness against radar noise, as weaker connections are compensated by stronger ones within the window, effectively smoothing the trajectory and correcting drift. This structure maintains a constant computational cost while ensuring an accurate enhanced odometry outcome.}
\end{itemize}

\revdg{\section{Related work}}

Radar-based SLAM solutions have seen a significant evolution, transitioning from basic 2D tracking applications to advanced 3D localization and mapping systems. Initially, radar was predominantly used in automotive contexts for 2D tracking on the horizontal (XY) plane to support systems such as advanced driver-assistance systems (ADAS), where the primary focus was on maintaining accurate vehicle positioning on roads and highways \cite{hong2020,adolfsson2021cfearRadarodometryconservative,article}. However, with the advent of 4D imaging radars, capable of capturing not only distance and velocity but also elevation information, radar-based SLAM has expanded its applicability to more complex three-dimensional (3D) environments \cite{4DRadarSLAM,article2}.

Existing techniques for radar-based odometry and Simultaneous Localization and Mapping (SLAM) primarily fall into three categories: filter-based methods, graph-based optimization, and neural network approaches. 

Traditionally, \textbf{filter-based methods}, such as the Extended Kalman Filters (EKF), have been utilized for recursive state estimation in radar-based SLAM\cite{9235254, callmer2011, 9470842}. These methods are effective at integrating data from multiple sensors and are particularly suited for real-time processing, as well as being robust against sensor noise. However, EKFs can struggle in highly dynamic or nonlinear environments, which can lead to performance degratation.

In response to the limitations of filter-based methods, \textbf{graph-based optimization }techniques, such as Pose Graph Optimization (PGO), have gained prominence \cite{shin2024multirobotrelativeposeestimation}. These methods are advantageous in handling the complex, nonlinear data typical of radar-based SLAM, allowing for the refinement of initial estimates and the correction of drift over time. PGO has proven especially effective for radar-based SLAM in large-scale environments, where maintaining global consistency is challenging. The performance of graph-based approaches, however, depends heavily on how the graph is constructed. For instance, in \cite{4DRadarSLAM}, relationships are formed between successive poses. However, due to the noisy nature of radar measurements, successive samples may not yield precise relationships for use in the optimizer, potentially leading to instability in the estimation. Our approach addresses this challenge by enhancing robustness against radar noise through a denser graph structure that establishes multiple interconnections between poses.

\textbf{Neural network} approaches have also been explored, leveraging deep learning to model complex relationships in radar data. For example, in \cite{safa2022fusingeventbasedcameraradar}, a SLAM architecture fuses an event-based camera with a Frequency Modulated Continuous Wave (FMCW) radar using a bio-inspired Spiking Neural Network (SNN) that learns features on the fly without requiring offline training. Similarly, in \cite{lu2020milliegosinglechipmmwaveradar}, milliEgo combines mmWave radar with an IMU using a deep neural network to improve pose estimation in challenging conditions like low-light or obstructed spaces. These methods can capture nonlinearities and have shown promise in radar-based odometry and SLAM. However, they often involve complex architectures and may not generalize well to unseen environments.

Additionally, \textbf{sensor fusion} is commonly employed across these methods to enhance performance by leveraging the strengths of different sensors. By combining radar data with other sensor modalities like cameras, IMUs, or thermal cameras, robustness is enhanced by compensating for the noise and sparsity of radar data with complementary information. For example, integrating inertial measurements from an IMU can improve the estimation of orientation and motion, aiding in the overall accuracy of the SLAM system. In \cite{10530463}, radar data is fused with thermal camera and IMU data, resulting in more reliable and accurate SLAM solutions in complex environments.
\revd{
Sensor fusion strategies are typically categorized as "loosely coupled" or "tightly coupled," depending on how the sensor data is integrated. In a tightly coupled approach, raw measurements from different sensors (e.g., IMU acceleration, radar Doppler, or even GNSS signals) are directly processed together in a unified estimation framework, often using filters or optimization-based techniques. This allows for deeper integration but increases computational complexity and dependency on accurate sensor models. In contrast, a loosely coupled approach processes each sensor independently to estimate intermediate states, such as velocity or pose, which are later combined at a higher level. This modular structure makes the system more adaptable and computationally efficient, at the cost of potentially weaker integration of constraints.
}
Here, we focus on the loose combination of IMU and radar, that is, a solution that integrates IMU and radar data without the need for a strict interdependence in the data fusion process, allowing each sensor to contribute independently to the vehicle's odometry estimation. Specifically, radar provides velocity estimations via the Doppler effect, while IMU data is used to track orientation and motion dynamics, which are then combined in the odometry framework. This approach balances robustness and computational efficiency, making it well-suited for real-time applications in challenging environments.

When working with radar, the estimation of ego-velocity is a key parameter employed in many algorithms, essential for accurate localization and mapping. Various techniques have been developed for this, such as in \cite{6728341}, where ego-velocity is estimated using a least-squares fitting method. Although this approach yields accurate results, radar measurements inherently carry noise due to factors such as multipath reflections, signal scattering, and limitations in angular resolution. This noise becomes especially pronounced in the Z-axis component, where lower vertical resolution and reduced field of view contribute to increased uncertainty and drift over time.

To mitigate these effects, some algorithms employ specific models to address drift in the vertical component. For instance, the EFEAR-4D approach \cite{wu2024efear4degovelocityfilteringefficient} estimates ego-velocity by assuming a non-holonomic vehicle model with movement constrained to a single axis, setting the vertical velocity to zero. This assumption simplifies the problem by modeling motion as circular in the XY plane, focusing solely on horizontal displacement and disregarding vertical shifts.

Other approaches, like that proposed in \cite{10100861}, address vertical drift by tightly coupling Doppler measurements with scan matching and integrating IMU and GNSS data through an iterative extended Kalman filter (iEKF). This integration allows the system to continuously correct vertical drift by leveraging Doppler velocity data alongside the IMU and GNSS information, improving robustness and positional consistency in large-scale environments.

To further improve on these methods and address radar noise in the z-axis, our proposed algorithm incorporates the vehicle’s motion model more comprehensively. By integrating IMU-provided orientation data to support ego-velocity estimation, this approach enhances accuracy and robustness, particularly in handling sensor noise, providing a more reliable estimation across all components, including the Z-axis.

Moreover, the creation of comprehensive datasets, such as NTU4DRadLM \cite{zhang2023ntu4dradlm4dradarcentricmultimodal} and others \cite{mscrad4r, boekema2024vodp}, has been pivotal in advancing radar-based SLAM research. These datasets offer a robust platform for evaluating SLAM algorithms across diverse structured and unstructured environments, enabling researchers to validate the effectiveness and reliability of radar-based approaches in real-world scenarios. By providing varied conditions and settings, these datasets help benchmark algorithmic performance, fostering the development of more resilient and adaptable SLAM systems.

\begin{figure}[t!] 
\centering
\includegraphics[width=0.8\textwidth]{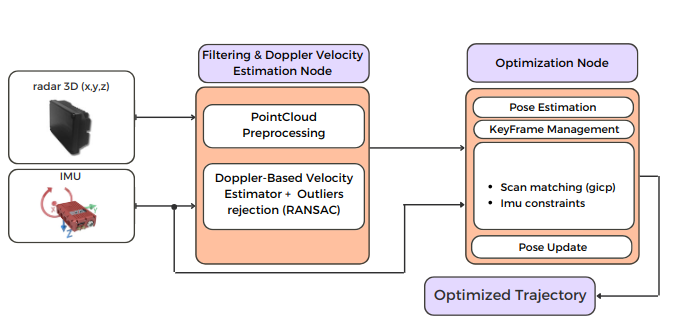} 
\caption{System Overview}
\label{fig:workflow}
\end{figure}

\section{Method overview}

The proposed odometry approach is shown in the flow diagram (Fig. \ref{fig:workflow}), it is divided into two main stages\revd{: preprocessing and velocity estimation, followed by optimization}. 
The first handles the calibration of data from the IMU and radar sensors, preprocessing, Doppler velocity estimation and filtering. It applies filtering techniques to the sensor readings and estimates the vehicle's ego-velocity using the Doppler effect at each radar point, taking into account the vehicle's degrees of freedom and reducing noise by excluding non-physically feasible movements. \revd{This method robustly filters out outliers by detecting inconsistent motion patterns, ensuring that only static elements contribute to the final velocity estimation. Furthermore, the IMU is an essential component of the velocity estimation process, as its orientation measurements directly constrain the system of equations used to compute the vehicle’s motion. These constraints help to reduce uncertainty, particularly in the vertical component, and improve the overall robustness of the estimation}.

\revd{The second stage focuses on the optimization process. The preprocessed radar and IMU data, along with the refined ego-velocity estimate, are used as input for initial pose estimation and graph construction. A pose graph is incrementally built, where radar scans and IMU readings define constraints that link consecutive keyframes. These constraints are formulated to enforce consistency in the estimated trajectory while accounting for the uncertainty inherent in radar data. The optimization framework then refines the vehicle’s trajectory by minimizing errors across the pose graph. To maintain computational efficiency, a sliding window approach is employed: Only the most recent \textit{n} keyframes are included in the optimization, and once poses exit this window, they are considered fixed and appended to the final trajectory without further modification. This ensures that the system remains scalable while preserving global consistency.}

\section{Velocity Estimation with Ground Vehicle Constraints}
\label{sec:vel}

Accurate velocity estimation is critical for autonomous navigation, especially when using radar-based odometry. The work presented in \cite{ITSC2013} introduces a method for estimating the velocity of a vehicle using Doppler radar data by measuring the projection of the vehicle's velocity vector $\mathbf{v}$ onto the line of sight of detected objects. Assuming static obstacles, the Doppler velocity $v_D$ for each of the obstacle points detected is defined as the projection of the vehicle's velocity vector $\mathbf{v}$ onto the unit vector $\hat{\mathbf{r}}$ pointing towards each radar-detected object as follows:.

\begin{equation}
    v_D = \hat{\mathbf{r}} \cdot \mathbf{v} = \hat{r}_x v_x + \hat{r}_y v_y + \hat{r}_z v_z
    \label{doppler}
\end{equation}

Each point can contribute to a system of linear equations involving the unknown velocity components $v_x$, $v_y$, and $v_z$, which can be solved using regular least squares. This method provides a reliable initial estimate of the vehicle's velocity by considering the contribution of multiple radar points in the environment:

\begin{equation}
\small
\begin{bmatrix}
v_{D,1} \\
v_{D,2} \\
\vdots \\
v_{D,N}
\end{bmatrix}
=
\begin{bmatrix}
\hat{r}_{1x} & \hat{r}_{1y} & \hat{r}_{1z} \\
\hat{r}_{2x} & \hat{r}_{2y} & \hat{r}_{2z} \\
\vdots & \vdots & \vdots \\
\hat{r}_{Nx} & \hat{r}_{Ny} & \hat{r}_{Nz}
\end{bmatrix}
\begin{bmatrix}
v_x \\
v_y \\
v_z
\end{bmatrix}
\label{dopplersystem}
\end{equation}

In order to have an accurate robot's velocity estimation from 4D-radar sensing, we need to account for those points that belong to moving objects. Otherwise, the velocity estimation will integrate such objects as noise. The usual approach is to build a robust estimator based on RANSAC \cite{ITSC2013}, detecting points belonging to moving objects and removing them from the estimator.
\revd{Despite these filtering steps, Doppler-based velocity estimation can still suffer from vertical drift and inconsistencies due to sensor noise. To improve robustness, we incorporate motion constraints based on IMU measurements, as detailed in the next section.}
\subsection{Improved Velocity Estimation Using Orientation}\label{egovel}
\revd{
The Doppler-based velocity estimation method provides an initial estimate by solving a system of linear equations derived from radar measurements. However, this estimation is susceptible to noise and inconsistencies, particularly in the vertical component.  To mitigate these issues, we incorporate motion constraints based on the vehicle's physical movement model, ensuring that only feasible velocities are considered. Ground robots exhibit structured motion patterns that can be leveraged to refine velocity estimation:}

    \begin{itemize}
        \item \revd{Their movement along the \textbf{Z-axis} is constrained by the vehicle's pitch and roll angles, meaning vertical velocity variations are primarily influenced by changes in orientation.}
    \end{itemize}
    \begin{itemize}
        \item \revd{Non-holonomic constraints restrict lateral motion, meaning that movement is largely constrained to the forward direction}
    \end{itemize}
\revd{
Since these constraints depend on the vehicle's orientation, the IMU plays a crucial role in providing real-time measurements. Using these properties, we reformulate the velocity estimation problem by incorporating orientation constraints.}
\revd{
Additionally, as in the basic Doppler model, a \textbf{RANSAC-based approach} is applied, which identifies as outliers those points with velocity different from the majority. Since most points in the environment correspond to static elements with a velocity that remains within a consistent range relative to the vehicle, dynamic objects exhibit atypical velocities that deviate significantly from this pattern. These outliers are filtered out to improve the accuracy of the estimation. Additionally, RANSAC helps mitigate the impact of \textbf{radioelectrical effects} such as \textbf{specular reflections} and \textbf{multipath effects}, which can also introduce inconsistent Doppler measurements.}

\subsubsection{Orientation Constraints for Holonomic Vehicles}\label{egovel}

\revd{For holonomic ground robots, motion primarily occurs along }the longitudinal \( V_x \) and lateral \( V_y \) axes. Instead of estimating all components independently, we can relate them through this change in orientation. The velocity components \( v_x \), \( v_y \), and \( v_z \) are therefore related as follows:

\begin{equation}
    \mathbf{v} =
    \begin{bmatrix}
    v_x \\
    v_y \\
    v_z
    \end{bmatrix} = 
    \begin{bmatrix}
    \cos(\theta)\, V_x\\
    \cos(\phi)\, V_y \\
    \sin(\theta)\, V_x + \sin(\phi)\, V_y
    \end{bmatrix}
    \label{velcomps}
\end{equation}

\vspace{0.1cm}

\noindent where \( \theta \) represents the pitch angle increment and \( \phi \) the roll angle increment, both from the IMU measurements, \revd{ corresponding to the orientation change between the current and previous radar scan}. In this way, we reduce the system of equations by one degree of freedom and, more importantly, constrain the value of \( \mathbf{v} \) so that it accounts for feasible robot velocities. Thus, if the vehicle is navigating over a relatively flat area, even though there may be noise in the measurement, the noise in \( v_z \) can be addressed and partially removed. \revd{Therefore, the system of equations to be solved, shown in \eqref{dopplersystem}, with the corresponding modification in the velocity vector seen in \eqref{velcomps}, becomes:}

\begin{equation}
\small
\begin{bmatrix}
v_{D,1} \\
v_{D,2} \\
\vdots \\
v_{D,N}
\end{bmatrix}
=
\begin{bmatrix}
\hat{r}_{1x} & \hat{r}_{1y} & \hat{r}_{1z} \\
\hat{r}_{2x} & \hat{r}_{2y} & \hat{r}_{2z} \\
\vdots & \vdots & \vdots \\
\hat{r}_{Nx} & \hat{r}_{Ny} & \hat{r}_{Nz}
\end{bmatrix}
\begin{bmatrix}
    \cos(\theta)\, V_x \\
    \cos(\phi)\, V_y \\
    \sin(\theta)\, V_x + \sin(\phi)\, V_y
\end{bmatrix}
\label{model1}
\end{equation}

\vspace{0.1cm}
\subsubsection{Orientation Constraints for Non-Holonomic Vehicles}

\revd{For non-holonomic vehicles, such as cars, motion is posible only in their longitudinal direction. This means that the velocity vector \( \mathbf{v} \), initially defined by it components \( v_x \), \( v_y \), and \( v_z \),  can be reduced to a single component, \( v_x \). To estimate the velocity in this scenario, we replace \( v_x \) with a vector of magnitude \( V \), wich can be represented using spherical coordinates. The resulting system of equation is then significantly constrained as shown in (\ref{model2}).}
\vspace{0.1cm}

\begin{equation}
\small
\begin{bmatrix}
v_{D,1} \\
v_{D,2} \\
\vdots \\
v_{D,N}
\end{bmatrix}
=
\begin{bmatrix}
\hat{r}_{1x} & \hat{r}_{1y} & \hat{r}_{1z} \\
\hat{r}_{2x} & \hat{r}_{2y} & \hat{r}_{2z} \\
\vdots & \vdots & \vdots \\
\hat{r}_{Nx} & \hat{r}_{Ny} & \hat{r}_{Nz}
\end{bmatrix}
\begin{bmatrix}
\sin\left(\dfrac{\pi}{2} - \theta\right) \cos\psi \\
\sin\left(\dfrac{\pi}{2} - \theta\right) \sin\psi \\
\sin\theta
\end{bmatrix}
V
\label{model2}
\end{equation}

\vspace{0.1cm}

\noindent Here,\revd{ $\psi$ corresponds to the yaw angle increment and $\theta$ to the pitch angle increment, both obtained from IMU measurements, corresponding to the orientation change between the current and previous radar scan}. This also reduces the number of unknowns to just one: the magnitude of the velocity $V$.

\section{Pose Graph-Based Odometry Algorithm}

The odometry algorithm developed in this work is designed to optimize the vehicle's trajectory by leveraging a set of constraints that refine the initial estimates provided by the Inertial Measurement Unit (IMU) and velocity measurements. For the IMU, pitch and roll angles are directly utilized as they are fully observable and generally precise. However, the yaw angle is prone to inaccuracies due to factors such as hard iron and soft iron distortions affecting the magnetometer. Consequently, yaw is integrated over time from angular velocity readings, which enhances the initial estimate but introduces cumulative drift. This drift is subsequently corrected by our Pose Graph-Based Odometry algorithm, which further refines the pose estimation.

The core of the algorithm involves creating and managing constraints that link different poses within an optimization window, constructing a local Graph-SLAM that optimizes the most recent \( n \) keyframes. These constraints are derived from the alignment of point clouds captured by the radar at different poses, IMU measurements, and the dynamics of the vehicle's movement. Preprocessing steps, such as downsampling using a VoxelGrid filter, are applied to reduce computational load and noise, thereby enhancing the robustness of the alignment process.

\subsection{Initial Estimation}
\label{egovel}

Accurate initial estimation of the vehicle's pose is essential, serving as the foundation for subsequent optimization steps. This estimation relies on the previous pose and a transformation matrix derived from the vehicle's ego velocity and orientation updates. A key challenge in this process is managing the discrepancy in data acquisition frequencies between the IMU and radar. The IMU typically operates at a significantly higher frequency than the radar, necessitating effective synchronization to ensure coherent pose estimation.

The transformation matrix representing the vehicle's current pose relative to the origin frame \( o \) is computed by combining the previous pose with the relative transformation derived from ego velocity and orientation updates:

\begin{equation}
    \mathbf{T}_{\text{current}}^{o} = \mathbf{T}_{\text{prev}}^{o} \cdot \mathbf{T}_{\text{curr}}^{\text{prev}}
    \label{transformation_matrix}
\end{equation}

Here, \( o \) denotes the origin reference frame, and \( \mathbf{T}_{\text{curr}}^{\text{prev}} \) is the relative transformation matrix between the current and previous poses. This relative transformation is constructed from the ego velocity \( \mathbf{v}_{\text{ego}} \) (as detailed in Section \ref{sec:vel}) and the orientation update quaternion \( \mathbf{q}_{\text{curr}}^{\text{prev}} \). Specifically, the translational component \( \mathbf{t}_{\text{curr}}^{\text{prev}} \) is computed from the linear velocity \( \mathbf{v}_{\text{ego}} \) and the time difference \( \Delta t \), while the rotational component \( \mathbf{q}_{\text{curr}}^{\text{prev}} \) is updated using the previous and current IMU orientation measurements.

\begin{align}
    \mathbf{t}_{\text{curr}}^{\text{prev}} &= \mathbf{v}_{\text{ego}} \cdot \Delta t \label{translation_component} \\
    \mathbf{q}_{\text{curr}}^{\text{prev}} &= \mathbf{q}_{\text{prev}}^{o^{-1}} \cdot \mathbf{q}_{\text{current}}^{o} \label{rotation_component}
\end{align}

\noindent Keyframes are systematically stored within a dedicated vector, capturing only those poses that exceed predefined thresholds for orientation (\( \delta q_{\text{rot}} \)) or translation (\( \delta t_{\text{trans}} \)). This selective inclusion ensures that the keyframe structure maintains a balance by focusing on significant changes in the vehicle's state. By limiting the number of keyframes, the optimization process avoids handling an excessive number of poses, which is essential for effective scan matching. Adequate separation between point clouds enhances alignment accuracy, while reducing the frequency of optimization operations ensures computational feasibility. Each keyframe encapsulates the current pose, the associated radar point cloud, and other pertinent data, which are subsequently integrated into the optimization process to improve the accuracy and robustness of trajectory estimation.

\subsection{GICP-based Alignment of RADAR point-clouds}
\label{gicp}
For each new keyframe, the algorithm aligns its corresponding point cloud with those of preceding keyframes within a sliding window. The Generalized Iterative Closest Point (GICP) algorithm \cite{7271006} is employed to refine the initial transformation estimate, which is derived from the relative pose differences between keyframes. The alignment process comprises the following steps:

\begin{enumerate}
    \item \textbf{Preprocessing}: Point clouds are downsampled using a VoxelGrid filter to reduce computational load and minimize noise.
    \item \textbf{Initial Guess}: An initial transformation estimate is computed based on the relative pose differences between the current keyframe and those within the optimization window.
    \item \textbf{Alignment}: The GICP algorithm iteratively refines the initial guess, resulting in a precise transformation that aligns the point clouds.
\end{enumerate}

\begin{figure}[t!]
\centering
\includegraphics[width=0.70\linewidth]{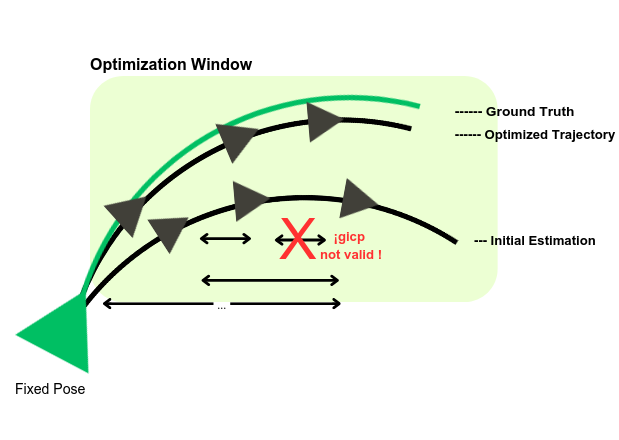}
\caption{Relationship between poses in the optimization window.}
\label{net}
\end{figure}

Upon successful convergence of the alignment, the final transformation is evaluated using the fitness score provided by the GICP algorithm. This score quantifies the quality of the alignment based on the average distance between corresponding points. A lower fitness score indicates a higher quality alignment, whereas a higher fitness score suggests potential unreliability. The fitness score is utilized to calculate a weight for the corresponding constraint, defined as:

\begin{equation}
\revd{w _{\text{GICP}}} = \frac{1.0}{2 f + \epsilon}
\end{equation}

\noindent where \revd{ \( w _{\text{GICP}}\) is the weight aplied to the corresponding cost function,} \( f \) represents the GICP fitness score and \( \epsilon \) is a small constant (e.g., \( 10^{-6} \)) to prevent division by zero.

This weight is applied to the GICP constraint, thereby attenuating the influence of constraints with higher fitness scores during the optimization process\revd{, while emphasizing those with lower fitness scores}. Additionally, alignments that yield fitness scores exceeding a predefined threshold \revd{$f_{\text{th}}$} are discarded, as they typically do not contribute to improving the initial pose estimation. Although such constraints receive a reduced weight via the factor \( w \), they are excluded to prevent potential degradation of the optimization outcome.

To enhance robustness, the odometry algorithm incorporates a strategy where poses are interconnected through multiple alignments within the sliding window (see Fig. \ref{net}). If a particular alignment fails to converge or produces a high fitness score, the associated constraint's weight is diminished, thereby minimizing its impact on trajectory optimization. However, due to the interconnected nature of poses within the mesh of alignments, successful alignments from other keyframes maintain the integrity of the optimization process.

This methodology ensures that the optimization process remains both accurate and computationally efficient by prioritizing reliable alignments and mitigating the influence of less reliable ones, thereby enhancing the overall robustness and precision of the vehicle’s trajectory estimation.

\subsection{Graph Construction and Optimization}
\label{sec:graph_optimization}

The pose graph optimization is formulated as a nonlinear least-squares problem, where the state vector \(\mathbf{X}\) is estimated by minimizing a set of cost functions, which define the residuals by relating the poses to each other and to external information. The state vector consists of individual position and orientation parameters for each pose within a sliding window of the last n poses. 

\begin{equation}
\mathbf{X} = [\mathbf{x}_1, \mathbf{x}_2, \mathbf{x}_3, ..., \mathbf{x}_n]
\end{equation}

\noindent where each pose $\mathbf{x}_i$ in the window is defined as \revdg{$\mathbf{x}_i=[x_i, y_i, z_i, q_{w_i}, q_{x_i}, q_{y_i}, q_{z_i}]$.} \revdg{To handle the non-Euclidean nature of quaternion-based orientation representations, the optimization is performed on a manifold. Specifically, quaternion updates are carried out in the local tangent space to ensure minimal and valid perturbations during optimization,  following common practice in odometry and state estimation problems.}

\revd{
The optimization process seeks to find the optimal state \(\mathbf{X}^*\) by minimizing the weighted sum of residuals of each cost function:  }

\begin{equation}
\mathbf{X}^* = \arg \min_{\mathbf{X}} \sum_{i} r_i^2
\end{equation}
\revd{
where \( r_i \) represents the residuals associated with different cost function.} 

\begin{figure}
    \centering
    \includegraphics[width=0.9\linewidth]{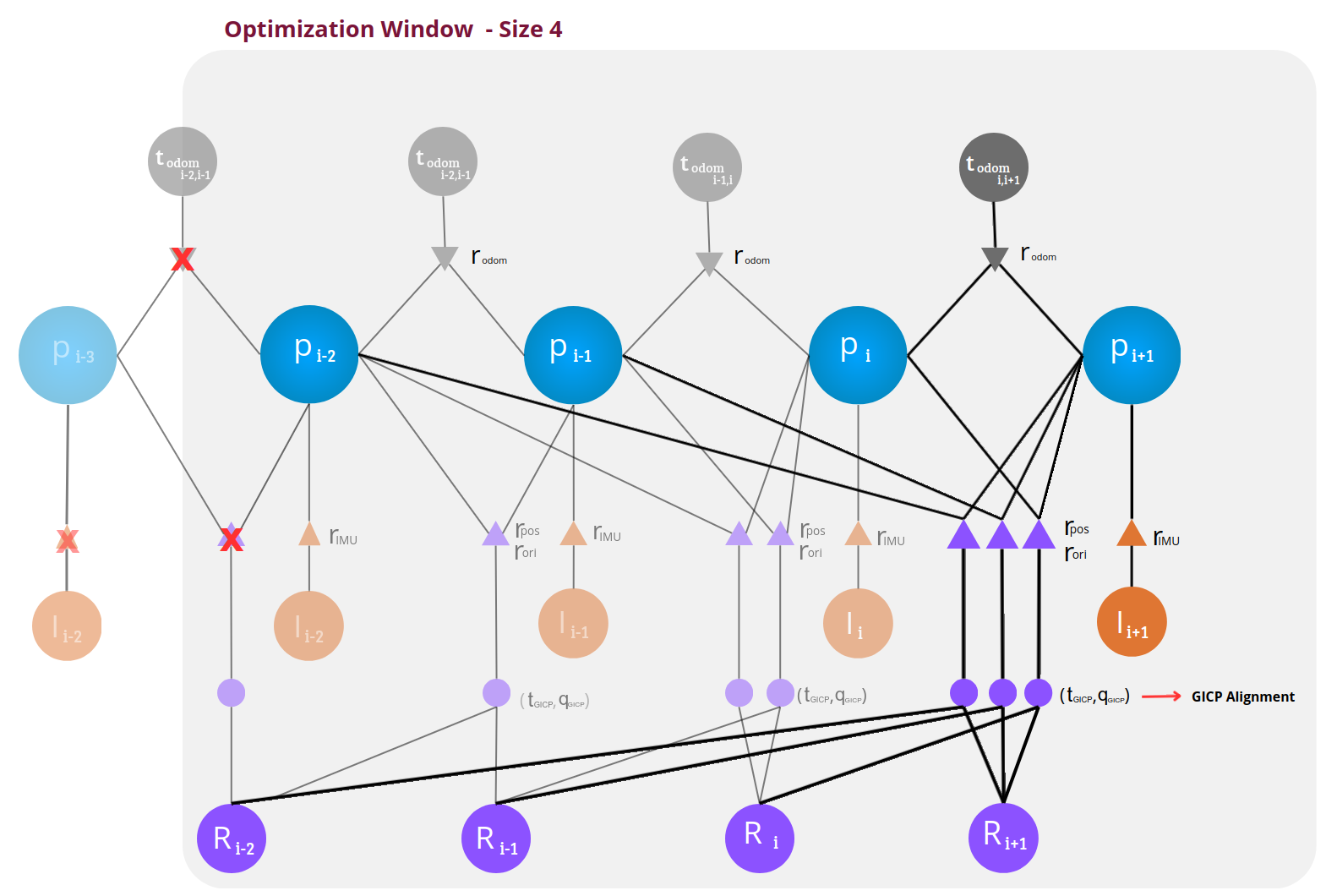}
    \caption{Optimization graph. The highlighted region shows new constraints added with the latest pose, while the shaded area represents existing constraints. Colors indicate radar measurements (purple), IMU measurements (orange), optimized poses (blue), and raw odometry transformations (gray).}
    \label{fig:graph}
\end{figure}

\revd{To efficiently solve this problem, we employ the Ceres Solver \cite{ceres-solver}, which optimizes the state vector by minimizing the cost functions within a sliding window encompassing the last n poses of the trajectory. Within this window, each pose is interconnected with the previous ones, forming an optimization graph, as shown in Figure \ref{fig:graph}, where each keyframe serves as a node with associated parameter blocks.}

\revd{As the trajectory progresses, older poses are removed from the optimization graph while new ones are incorporated, ensuring a balance between global consistency and adaptability to new sensor data. This sliding-window approach allows recent observations to have a stronger influence on the estimated trajectory while still maintaining coherence with past constraints.}

\revd{The optimizer is configured to use a Levenberg-Marquardt trust-region strategy combined with a sparse normal Cholesky solver to balance computational cost and accuracy. Additionally, a Tukey robust kernel is applied to mitigate the influence of outliers, particularly those arising from noisy sensor measurements. The optimization runs for a maximum of 100 iterations, with convergence criteria set to their default values (\texttt{function\_tolerance}: \(10^{-6}\), \texttt{gradient\_tolerance}: \(10^{-10}\)).
}

\subsubsection{Cost Function Definitions}

Two primary cost functions derived from the GICP constraint are applied, where each function enforces alignment between successive \revd{keyframe radar measurements} in the window, thus refining both position and orientation estimates across the trajectory. \revd{ As shown in Figure \ref{fig:graph}, this constraint, marked by a purple triangle, links radar measurements to provide a refined transformation between poses, as detailed in Section \ref{gicp}.}

\begin{itemize}
  \item \textbf{Position Cost Function}: This function quantifies the residuals by evaluating the discrepancy between the estimated translation vector and that derived from GICP. The translation error \( r_{\text{pos}} \) is expressed as:
    
    \begin{equation}
    r_{\text{pos}} = w_{p} \cdot  w _{\text{GICP}}  \cdot \left( t_{\text{GICP}} - q_{a}^{o^{-1}} \left( t_{b}^{o} - t_{a}^{o} \right) \right)
    \end{equation}
    
    where \( t_{a}^{o} \) and \( t_{b}^{o} \) represent the translation vectors of poses \( a \) and \( b \), respectively, \revd{and \( q_{a}^{o} \) the global quaternion orientation of pose a}. The weight \revd{\( w _{\text{GICP}}\)} is directly applied to the residual to modulate its influence based on the fitness score as shown in \ref{gicp}, ensuring that less reliable constraints yield a diminished impact on the final optimization solution.
    
    \vspace{0.1cm}

    \item \textbf{Orientation Cost Function}: This function quantifies the residuals by evaluating the discrepancy  between the GICP-derived quaternion and the estimated quaternion. The rotational error \(\ {r}_{\text{ori}}\) is given by:
    
    \begin{equation}
    \ {r}_{\text{ori}} =  \ {w}_{o} \cdot w _{\text{GICP}} \cdot \left( (\ {q}_{\text{GICP}}^{-1} \cdot \ {q}_{a}^{o^{-1}} \cdot \ {q}_{b}^{o}) - q_I\right)
    \end{equation}
    
     where \( q_{a}^{o} \) and \( q_{b}^{o} \) are the \revd{global} orientation quaternions of poses \( a \) and \( b \), respectively, and \( q_{\text{GICP}} \) refers to the relative rotation estimated by the GICP algorithm between the two poses. Here, \( q_I \) denotes the identity quaternion\revd{, ensuring that the rotational residual is zero when no rotation is present.} Analogous to the position cost function, the weight \revd{\( w _{\text{GICP}}\)}
     is applied directly to the residual to adjust its impact in the optimization process, modulating its influence in the optimization to favor constraints with higher reliability.
\end{itemize}

Two weights, \(\ {w}_{p}\) and \(\ {w}_{o}\) are multiplied in both previous cost functions to normalize and balance the weighting scales, particularly to account for parameters prone to higher noise. A Tukey Robust Kernel function is also applied to each cost function, effectively limiting the influence of outlier residuals with disproportionately large values on the optimization outcome.

Additionally, the following two constraints are also incorporated into the optimization process:

\begin{itemize}
    \item \textbf{IMU Pitch and Roll \revd{Cost Function}:} This cost function, \revd{shown in Figure \ref{fig:graph}  as orange elements,} enforces consistency for the pitch and roll angles provided by the IMU, remain consistent throughout the optimization. Since the IMU-derived pitch and roll angles typically exhibit bounded error, this constraint helps maintain the poses within an acceptable range, enhancing the stability of the pose estimates. \revdg{These values correspond to the orientation estimates provided directly by the commercial IMUs, which typically includes onboard EKF filtering to fuse gyroscope, accelerometer, and magnetometer data. This process compensates for sensor biases and provides stable roll and pitch angles, usually with errors below \(0.5^\circ\). In contrast, yaw, being more affected by magnetic perturbations, is handled indirectly through the alignment-based constraints in the graph.}
    
    \begin{equation}
    \ {r}_{\text{IMU}} =  (\ {q}_{\text{IMU}}^{-1} \cdot \ {q}_{a}^{o^{-1}}) - q_I
    \end{equation}
where both \( q_{\text{IMU}} \) and \( q_{a} \) are orientation quaternions constructed by setting the \textit{yaw} component to zero, thus considering only rotations in the \textit{pitch} and \textit{roll} angles. \( q_{\text{IMU}} \) is derived from the IMU data, while \( q_I \) serves as the identity quaternion. This cost function ensures that, even when the optimizer adjusts other cost functions, it avoids significant deviations in the pitch and roll angles from the values provided by the IMU, preserving consistency.

This constraint is applied locally to each keyframe within the optimization window and is weighted adaptively based on the reliability of the GICP-derived orientation constraint. Specifically, when the GICP constraint has a lower weight, indicating reduced reliability, the IMU constraint exerts a stronger influence to maintain pitch and roll consistency. In contrast, when the GICP constraint is deemed reliable (i.e., with a high fitness score), the IMU constraint’s weight is reduced, allowing for subtle adjustments to pitch and roll that improve alignment.

    \vspace{0.1cm}
    
\item \textbf{Vehicle Dynamics \revd{Cost Function}:} To reflect the ground vehicle's physical limitations, this constraint\revd{, shown in Figure \ref{fig:graph} (gray elements),} restricts vertical (Z-axis) movements inconsistent with the vehicle’s dynamics. This is particularly advantageous given that GICP often introduces noise into the z-component, occasionally resulting in unrealistic vertical displacements. By imposing this constraint, the optimization better aligns with the physical motion constraints of ground vehicles, reducing the likelihood of non-physical z-axis shifts.

\begin{equation}
r_{\text{dyn}} = \left( t_{\text{odom}} - q_{a}^{o^{-1}} (t_{b}^{o} - t_{a}^{o}) \right)
\end{equation}

where \( t_{\text{odom}} \) denotes the z-axis translation between keyframes \( a \) and \( b \), referenced in the coordinate frame of pose \( a \). This translation measurement is derived from direct odometry and limits excessive z-axis displacements that may arise from GICP-induced noise.
\revd{This constraint incorporates the effect of the refined velocity estimation shown in Section \ref{sec:vel} into the optimization process. Since the initial odometry guess is directly influenced by the velocity refinement, the enforced limitation on vertical displacement ensures that the optimization process remains consistent with the improved motion estimate. As a result, even when GICP introduces noise in the z-component, the optimization process remains consistent with the physically valid motion dictated by the velocity-based odometry, preventing unrealistic vertical shifts while maintaining consistency with the vehicle's dynamics. This constraint could also be extended to non-holonomic ground vehicles by incorporating the lateral y-component into the residual. However, it has been applied specifically to the z-component to simplify the optimization problem, as GICP primarily intoduces errors in the vertical direction.}
    
\end{itemize}

The solver prioritizes GICP constraints when they exhibit high reliability, ensuring that the trajectory estimation relies primarily on accurate geometric alignments. However, as the reliability of the GICP constraints decreases, the IMU and vehicle dynamics constraints assume a greater role, ensuring consistent and physically plausible pose adjustments. By dynamically adjusting constraint weights based on their reliability, the optimization balances accuracy and robustness, maintaining computational efficiency within a fixed budget.

\section{Experimental Results}\label{AA}

To validate the results of our algorithm, we employed the NTU4DRadLM dataset provided by \cite{zhang2023ntu4dradlm4dradarcentricmultimodal}. This dataset includes four different trajectories. The first two trajectories, which cover a shorter distance (246 and 1017 m.), are executed with a handcart at an approximate speed of 1 m/s. In these scenarios, it is important to consider that the vehicle is a holonomic ground vehicle, meaning it can move independently in its x and y components, while independent movement in the z component is restricted. Therefore, the proposed approach in \eqref{model1} is used. In the remaining trajectories, the distance covered is longer (4.79 and 4.23 Km.) with speeds of 25 and 30 m/s, respectively. As the vehicle used in these cases is a car (non-holonomic), we thereby apply the proposed approach in \eqref{model2}, which only allows for the vehicle's longitudinal movement.

We first demonstrate how considering the vehicle's motion model affects direct odometry by applying its corresponding system of equations, followed by the evaluation of the proposed optimization algorithm.

To evaluate the performance of our algorithm, we used the \texttt{rpg-trajectory-evaluation} package \cite{zhang2018tutorial}, which is the same tool employed by the radar SLAM algorithm \cite{4DRadarSLAM}, with which we will later compare. This package provides key metrics such as RMSE (Root Mean Square Error) values for relative translation  \(t_{\text{rel}}(\%)\), relative rotation  \(r_{\text{rel}}(\text{deg/m})\), and absolute translation \(t_{\text{abs}}(\text{m})\). 

In the first part, when analyzing direct odometry, the absolute error in the x, y, and z components is calculated separately. Although the \texttt{rpg-trajectory-evaluation} package uses Euclidean distance to compute the error, this method makes the variations in the z-axis less noticeable due to the typically larger errors in the x and y components of non-optimized trajectories. Therefore, the absolute error for each component is computed independently.

\subsection{Impact of Velocity Estimation Model Adjustments for Ground Vehicles}\label{ego-vel results}
\begin{figure}[t!]
\centering
\begin{minipage}{\linewidth}
    \centering
    \includegraphics[width=0.9\linewidth]{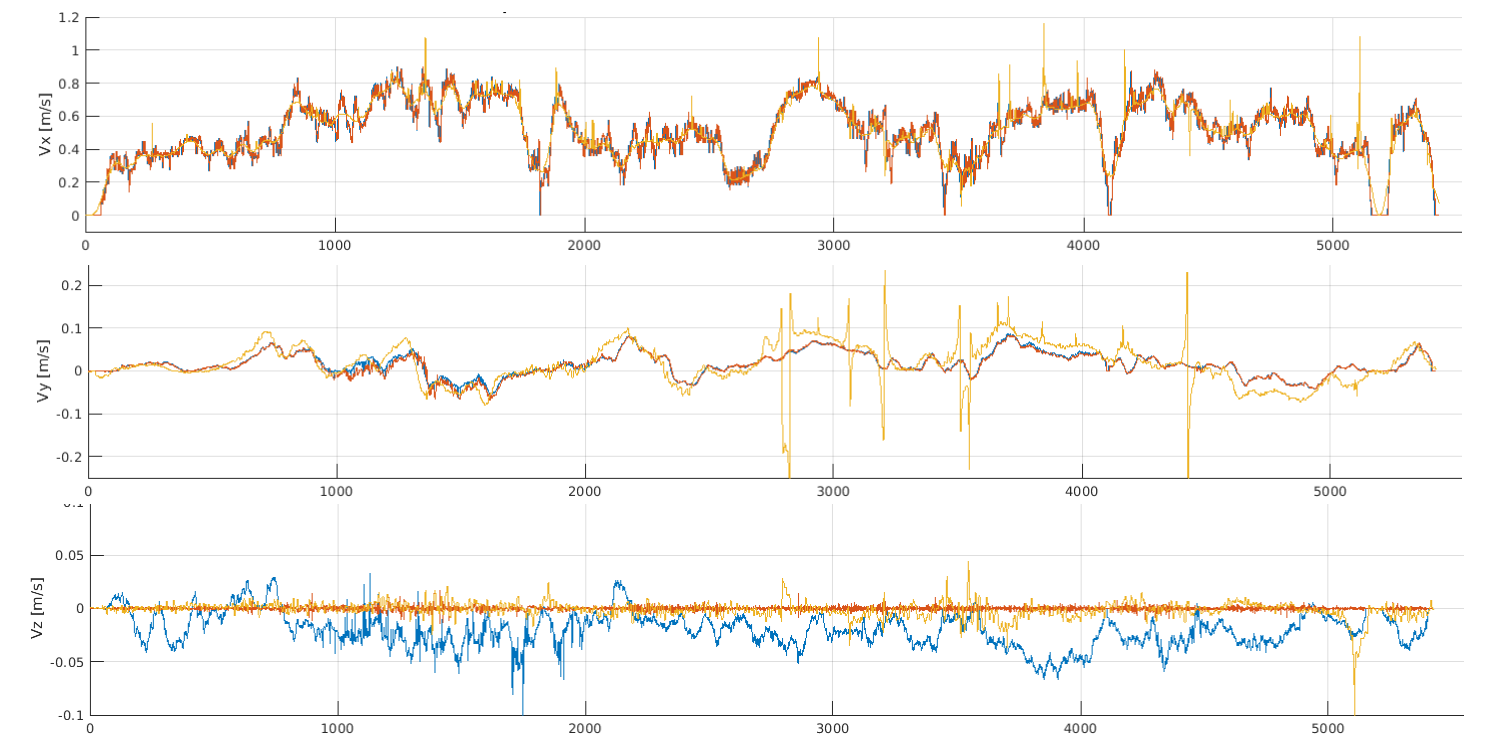}
\end{minipage}

\vspace{5mm} 

\begin{minipage}{\linewidth}
    \centering
    \includegraphics[width=0.9\linewidth]{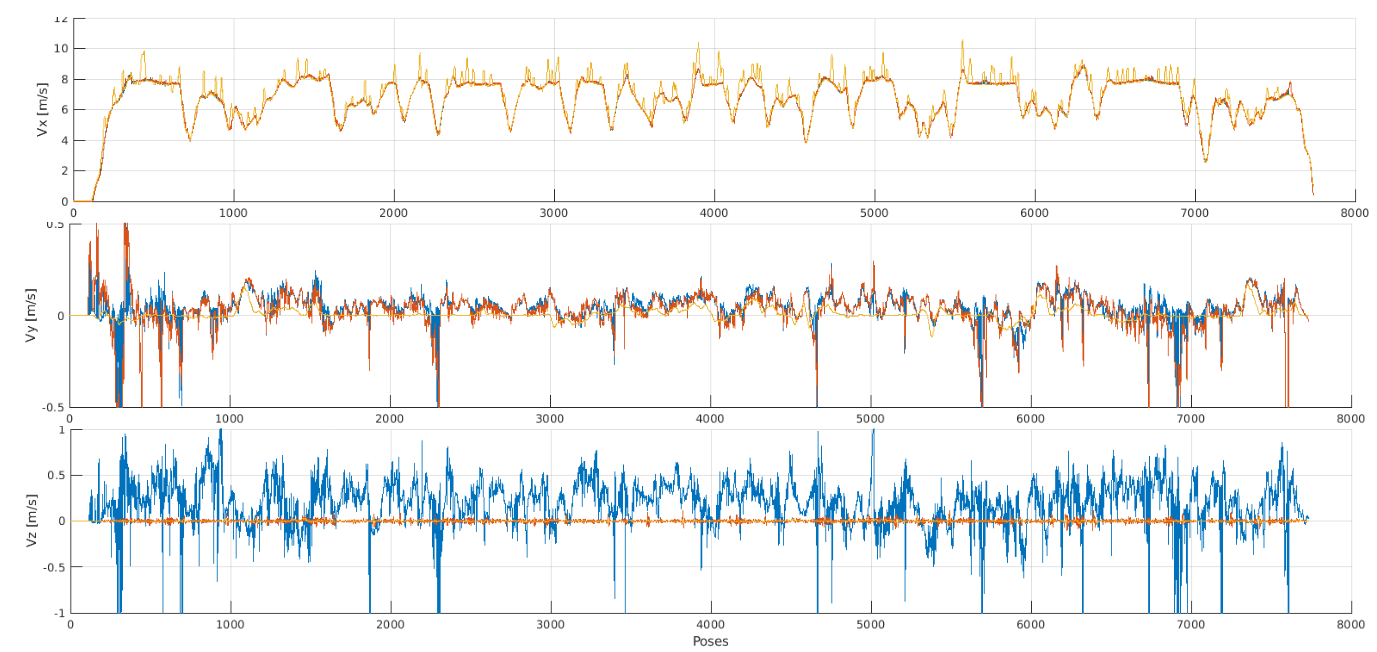}
\end{minipage}

\caption{Velocity component comparisons for the maps "cp" and "loop1". The blue curve corresponds to the \textit{Classic Velocity Model}, the orange curve denotes the velocity estimation from \textit{Our Proposed Method}, and the yellow curve represents the ground truth velocity, obtained by considering the vehicle's dynamics.}
\label{velcomp}
\end{figure}

This section presents the improvements in vehicle velocity estimation achieved through modifications to the classical model equations introduced in Section \ref{egovel}. These adjustments specifically enhance the linear velocity estimates in the vehicle's local frame. Figure \ref{velcomp} shows the velocity components \( V_x \), \( V_y \), and \( V_z \) for both the classical velocity estimation model and our proposed model. Additionally, the ground truth linear velocity was estimated by incorporating the vehicle's motion model for each map pair.

Our model achieves a closer approximation to the ground truth, particularly in the vertical component \( V_z \), where the most notable improvement is observed, reducing the unintended vertical drift introduced by noise in the classical model. For the handcart vehicle, the vertical component displays higher noise levels, as expected, due to its less precise suspension system, making it more susceptible to road imperfections and bumps.

\begin{figure}[t!]
    \centering
    \begin{minipage}{0.4\textwidth}
        \raggedright
        \includegraphics[width=0.4\textwidth]{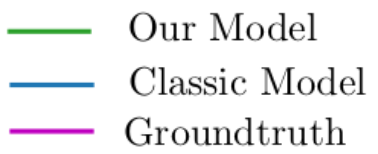}\\ 
        \centering
        \includegraphics[width=1.0\textwidth]{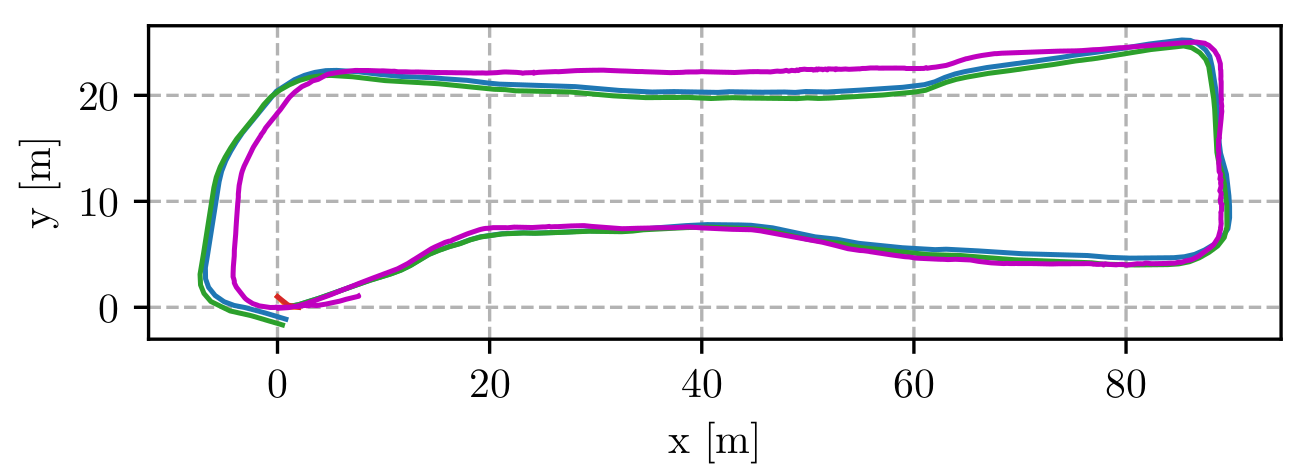} \\
        \includegraphics[width=1.0\textwidth]{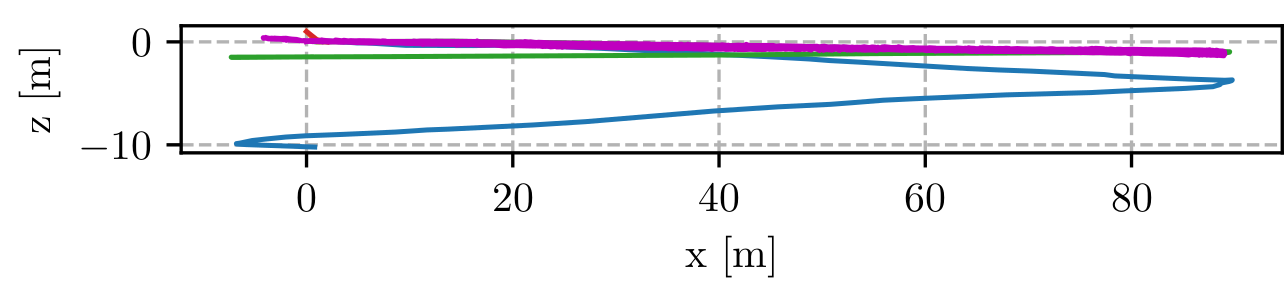}\\
       
    \end{minipage}
    \begin{minipage}{0.4\textwidth}
        \centering
        \includegraphics[width=0.9\textwidth]{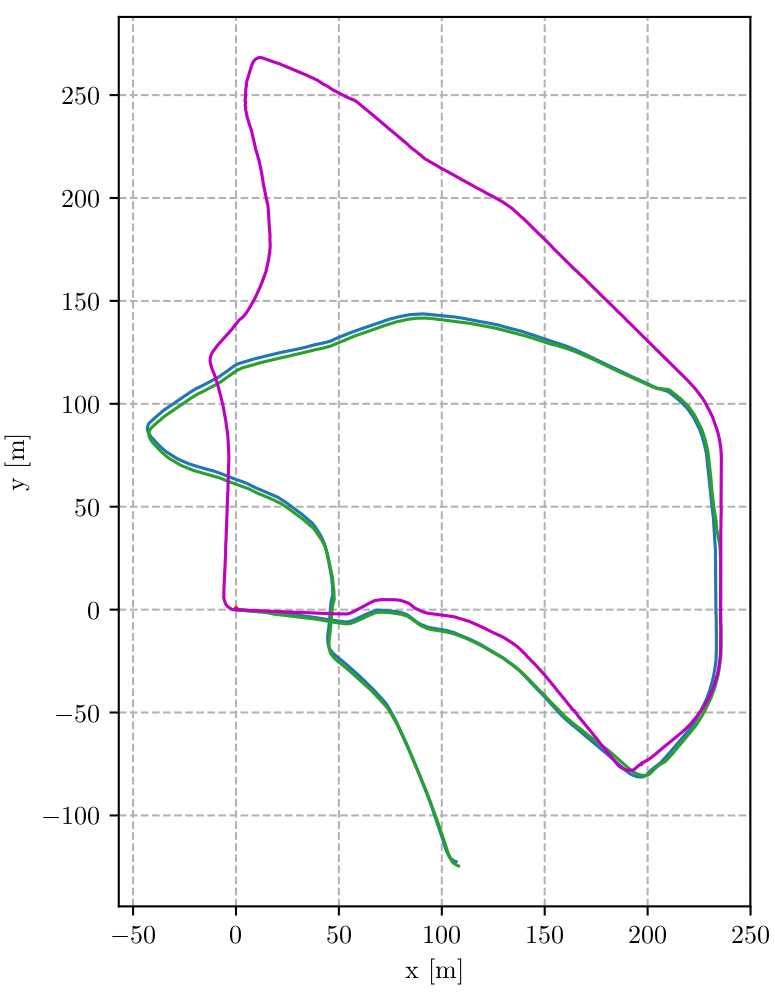} \\
        \includegraphics[width=0.9\textwidth]{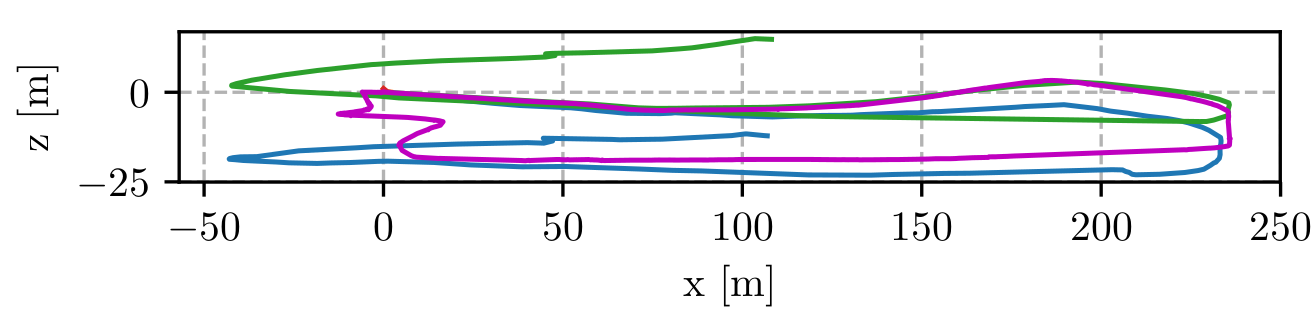}
    \end{minipage}\\
    \vfill
    \begin{minipage}{0.4\textwidth}
        \centering
        \includegraphics[width=1.0\textwidth]{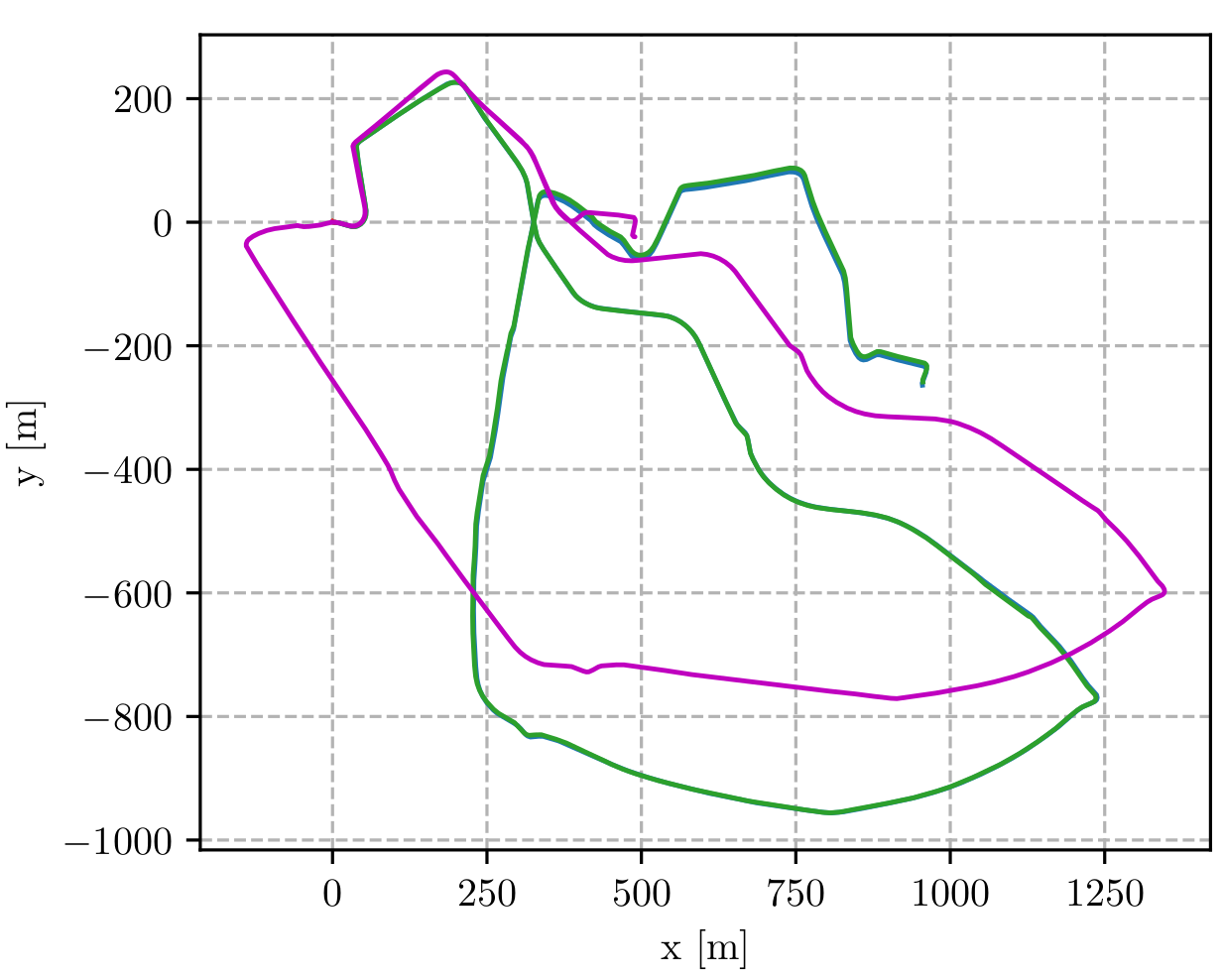} \\
        \includegraphics[width=1.0\textwidth]{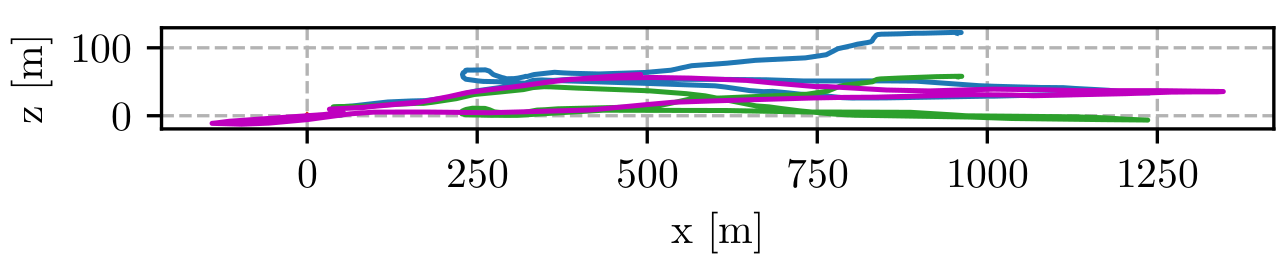}
    \end{minipage}
    \begin{minipage}{0.4\textwidth}
        \centering
        \includegraphics[width=0.9\textwidth]{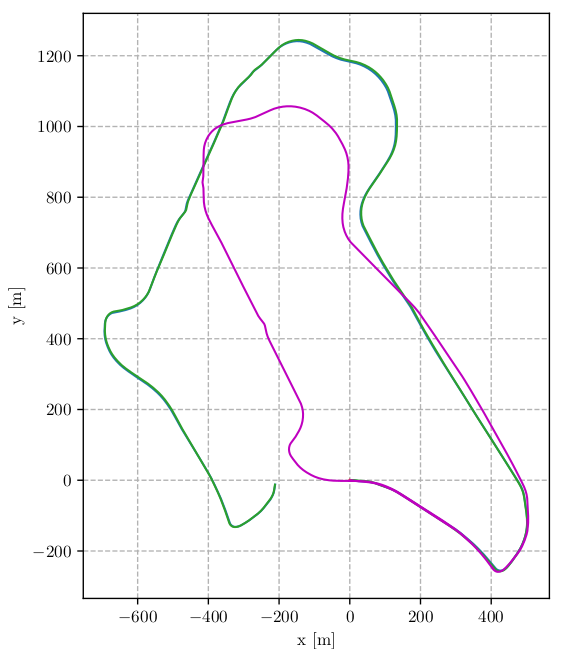} \\
        \includegraphics[width=0.9\textwidth]{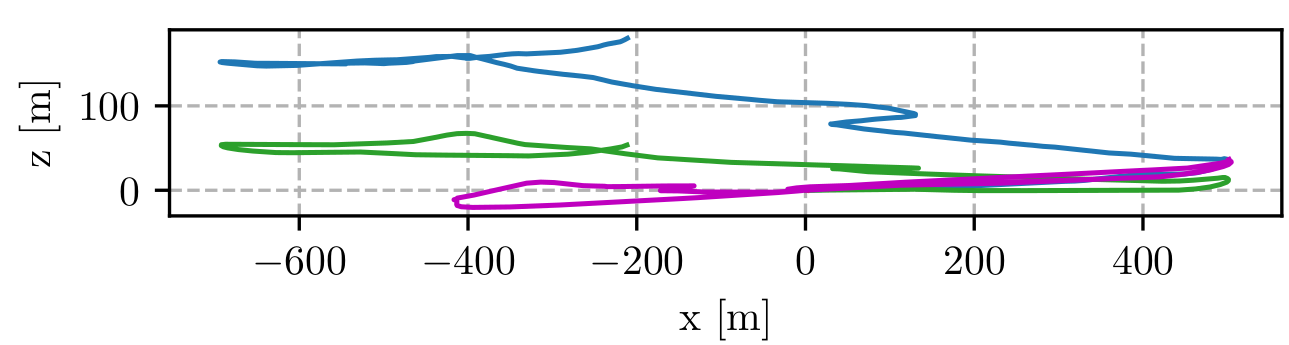}
    \end{minipage}
    
    \caption{Top and Side views of four different maps, arranged to optimize visual space.}
    \label{maps}
\end{figure}

To evaluate the impact of both velocity models on trajectory estimation, we estimated trajectories using both models across all four maps, as shown in Figure \ref{maps}. For the first two maps, we applied the velocity model of a holonomic ground vehicle, while for the latter two maps, we used the model for a non-holonomic vehicle.
 While both models (the classical model and our proposed model)  produced similar trajectories in the \( xy \)-plane, they exhibited notable differences along the \( z \)-axis due to noise primarily affecting the vertical component. This noise introduces a cumulative bias that accumulates with each odometry step, leading to unintended vertical drift inconsistent with the vehicle's motion constraints.

By mitigating this error in our proposed model, the trajectory better aligns with the ground truth, with the vertical component reflecting genuine vertical displacements rather than accumulated drift. For the "cp" and "loop1" maps, as illustrated in Fig. \ref{velcomp}, the holonomic vehicle exhibits a downward drift due to a consistent negative noise in the vertical component, whereas the non-holonomic vehicle shows an upward drift resulting from a positive noise bias, as observed in the \( V_z \) component of the velocity plots in Fig. \ref{velcomp}.

In addition to the velocity profiles, Table \ref{Table01} presents the absolute translational errors for each trajectory. The \( t_{\text{abs}, x} \) and \( t_{\text{abs}, y} \) errors remain similar for both models, whereas a significant reduction in \( t_{\text{abs}, z} \) error is observed in most cases, highlighting the effectiveness of our proposed model in reducing vertical drift.

\begin{table}[t!]
\centering
\footnotesize
\setlength{\tabcolsep}{4pt}
\begin{tabular}{lccc|ccc}
\textbf{} & \multicolumn{3}{c}{\textbf{Cp}} & \multicolumn{3}{c}{\textbf{Nyl}} \\
\cmidrule(lr){2-4} \cmidrule(lr){5-7}
& $t_{\text{abs}, x}$ & $t_{\text{abs}, y}$ & $t_{\text{abs}, z}$ & $t_{\text{abs}, x}$ & $t_{\text{abs}, y}$ & $t_{\text{abs}, z}$ \\
\midrule
\textbf{Classic} & 1.209 & 0.710 & 4.344 & 27.329 & 60.901 & \textbf{4.898} \\
\textbf{Ours}    & 1.383 & 0.810 & \textbf{0.485} & 27.388 & 62.508 & 9.322 \\
\\
\textbf{} & \multicolumn{3}{c}{\textbf{Loop1}} & \multicolumn{3}{c}{\textbf{Loop2}} \\
\cmidrule(lr){2-4} \cmidrule(lr){5-7}
& $t_{\text{abs}, x}$ & $t_{\text{abs}, y}$ & $t_{\text{abs}, z}$ & $t_{\text{abs}, x}$ & $t_{\text{abs}, y}$ & $t_{\text{abs}, z}$ \\
\midrule
\textbf{Classic} & 193.907 & 106.420 & 32.031 & 229.866 & 137.535 & 90.738 \\
\textbf{Ours}    & 194.972 & 105.847 & \textbf{14.076} & 231.319 & 140.468 & \textbf{16.313} \\
\bottomrule
\end{tabular}
\caption{Absolute errors in $t_{\text{abs}, x}$, $t_{\text{abs}, y}$, and $t_{\text{abs}, z}$ components (in meters) across different datasets for both models.}
\label{Table01}
\end{table}

Therefore, by implementing these improvements to reduce the impact of noise on velocity, the vehicle can follow a more reliable trajectory, relying more closely on IMU data to update its position while filtering out lateral or vertical movements that are not physically feasible. Nevertheless, it remains susceptible to consistent IMU errors, especially on maps like 'nyl,' where ongoing IMU inaccuracies lead to discrepancies that complicate interpretation. This limitation is primarily due to input data quality rather than the model adjustments.

\subsection{PoseSLAM optimization results}\label{Optimized results}

The proposed algorithm was tested on the four different scenarios using the following parameters: a translation threshold of $\delta t_{\text{trans}} = 1.0$ meters, a rotation threshold of $\delta q_{\text{rot}} = 5^\circ$, and a \revd{fitness score threshold of $f_{\text{th}}= 3.5$}. The weight matrices were set as follows: $w_o$ was kept as $[1.0, 1.0, 1.0,1.0]$, without influencing the weight of the residual, while $w_p$ was defined as $[0.1, 0.1, 0.1]$ to scale the components based on the vehicle’s orientation.

\begin{figure}[t!]
    \centering
    \begin{minipage}{0.5\textwidth}
        \raggedright
        \includegraphics[width=0.3\textwidth]{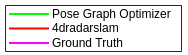}\\ 
        \vspace{1.0cm}
        
        \centering
        \includegraphics[width=0.9\textwidth]{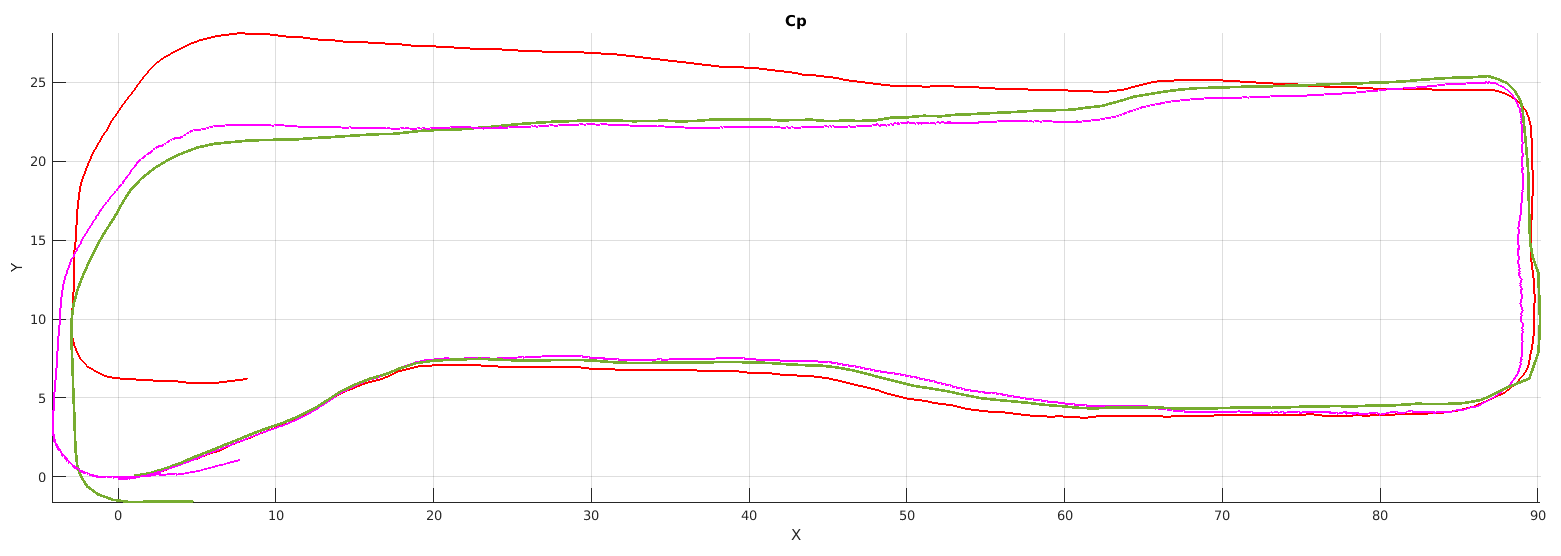} \\
        \includegraphics[width=0.9\textwidth]{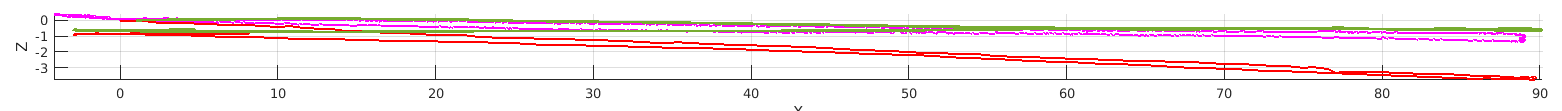}\\
       
    \end{minipage}
    \hfill
    \begin{minipage}{0.45\textwidth}
        \centering
        \includegraphics[width=0.6\textwidth]{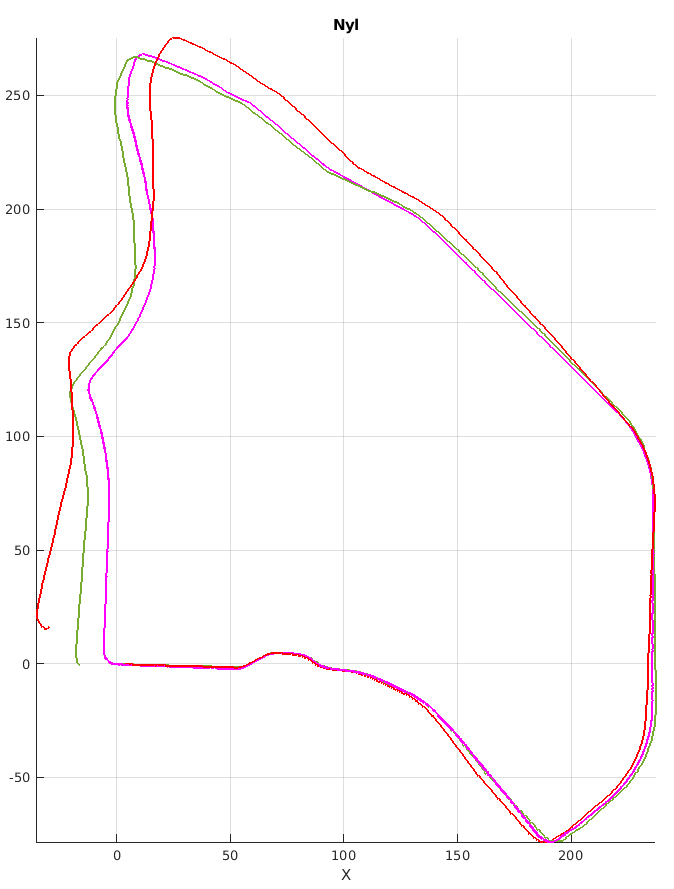} \\
        \includegraphics[width=0.60\textwidth]{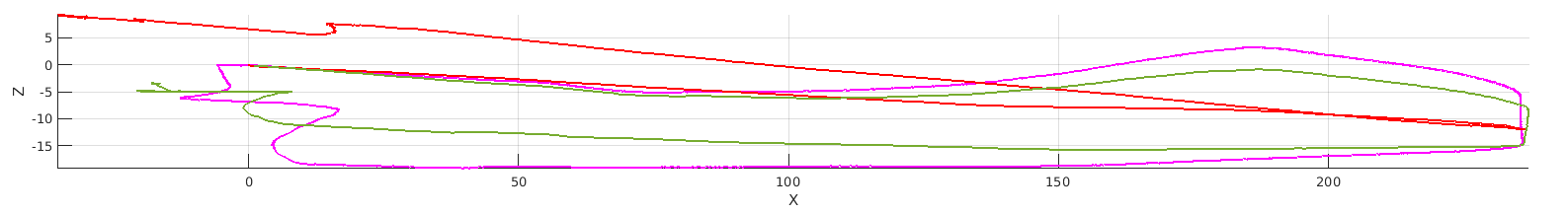}
    \end{minipage}\\
    \vspace{0.5cm}
    \begin{minipage}{0.45\textwidth}
        \centering
        \includegraphics[width=0.9\textwidth]{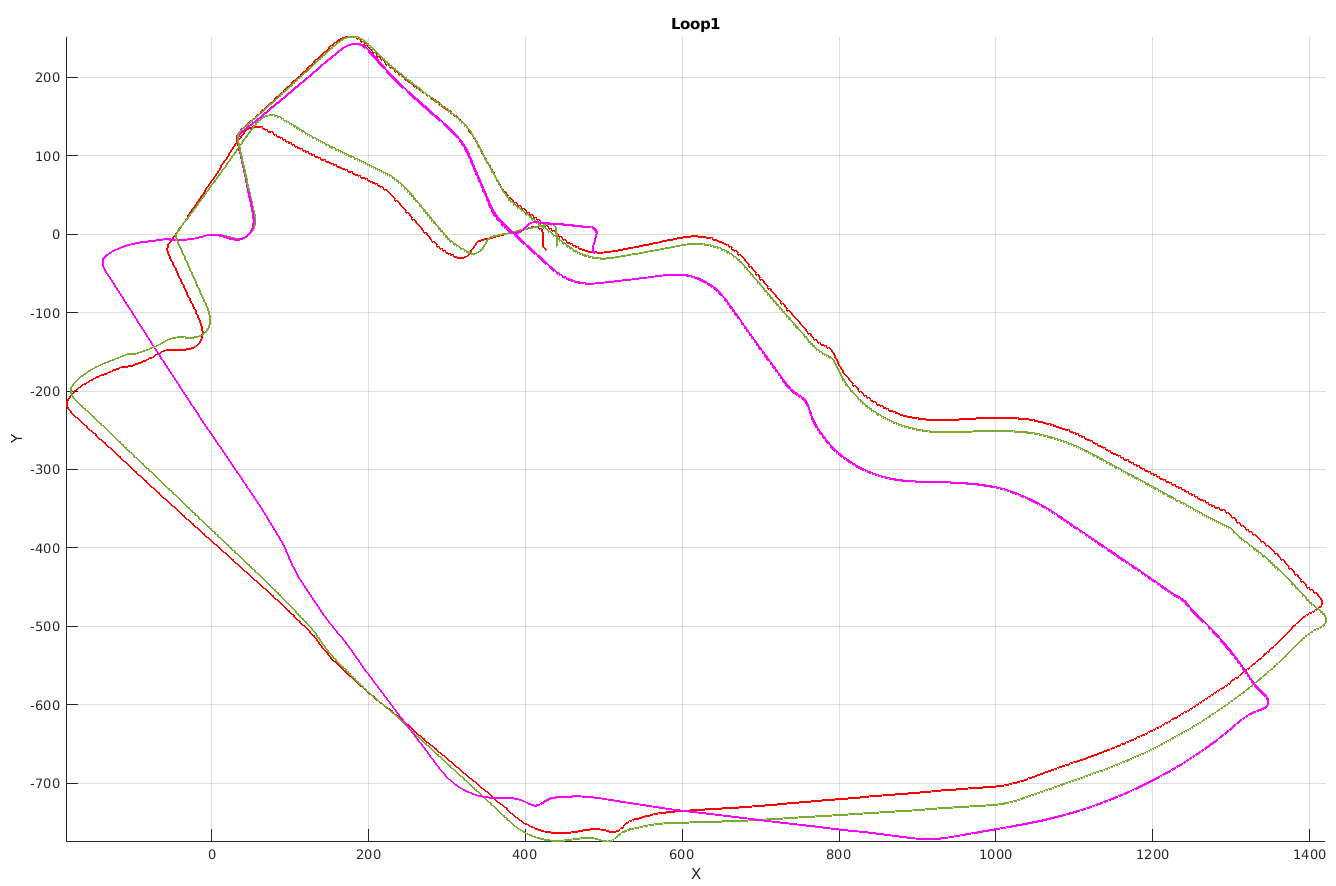} \\
        \includegraphics[width=0.9\textwidth]{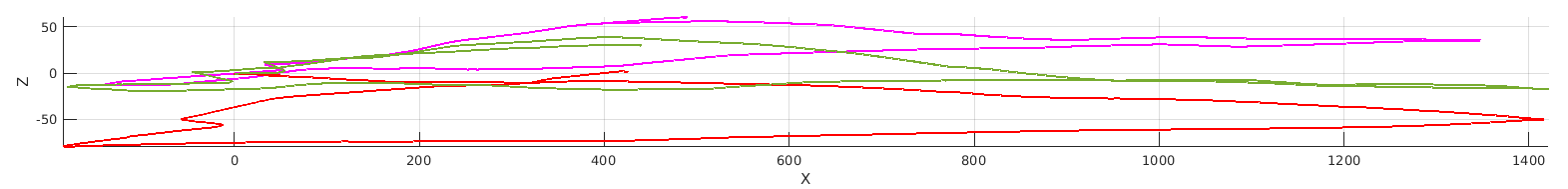}
    \end{minipage}
    \hfill
    \begin{minipage}{0.5\textwidth}
        \centering
        \includegraphics[width=0.60\textwidth]{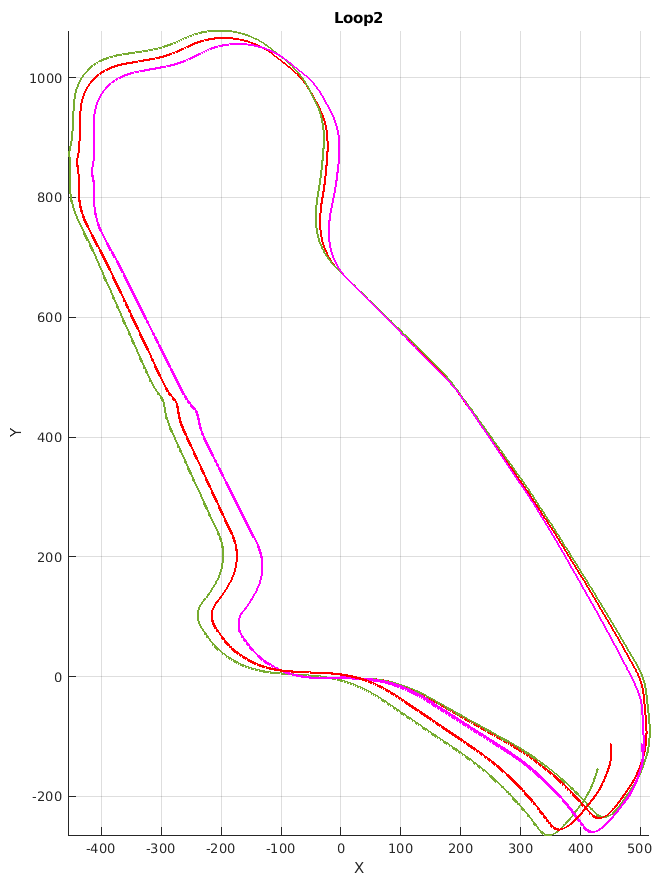} \\
        \includegraphics[width=0.60\textwidth]{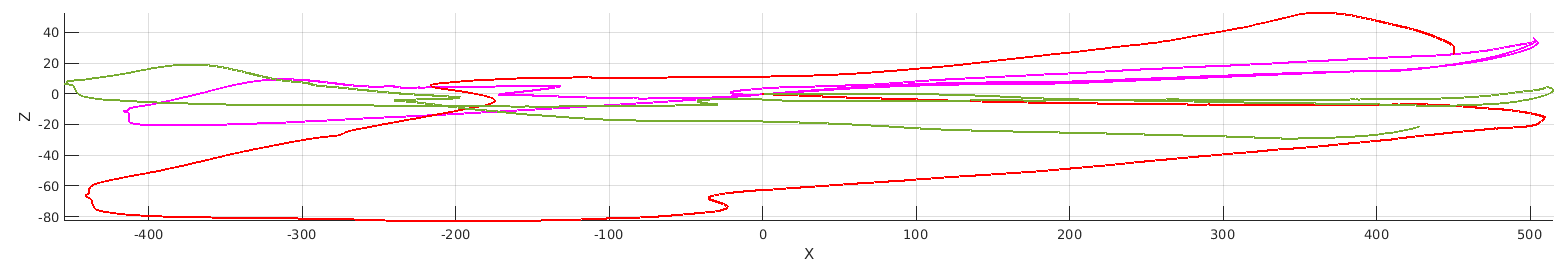}
    \end{minipage}
    
     \caption{\revdg{Optimization results on 4 different maps: 'Cp', 'nyl','loop1' and 'loop2'.}}
    \label{fig:optmaps}
\end{figure}

The results of the optimization are presented in Fig. \ref{fig:optmaps}, where both top and side views are shown for a comprehensive comparison of trajectories. Table \ref{Table02}  provides metrics for both direct and optimized odometry along with the variations of 4DRadarSLAM \cite{4DRadarSLAM} (GICP and ADPGICP, which incorporates a probabilistic distribution into the GICP algorithm, with and without loop closure). \revdg{We focus primarily on absolute error, highlighted in bold in Table \ref{Table02}, as it is the most critical metric, though relative errors are also provided. In our case, our approach does not include large loop closure, but it may be of interest to compare all metrics to assess how far they are from the loop closure algorithm also proposed in \cite{4DRadarSLAM}.}

When comparing our optimized trajectory with the direct one, it is evident from the results that the optimization notably reduces the drift in alignment with the ground truth. 
\revdg{Comparison with previous results from \cite{4DRadarSLAM} shows that our method achieves improved performance, with clear reductions in absolute error across all maps, while relative translational and rotational errors remain similar. It should be noted that loop closure algorithms are shown here for reference but are not directly compared, as they entail an additional optimization step not implemented in pure odometric methods. That said, our approach also provides absolute errors very similar to the loop closure methods, and even outperforms them on certain trajectories.}

\begin{table}[t!]
\renewcommand{\arraystretch}{1.0}
\resizebox{\textwidth}{!}{%
\centering
\begin{tabular}{lccc ccc ccc ccc}
\cmidrule(lr){2-13}
& \multicolumn{3}{c}{\textbf{cp}} & \multicolumn{3}{c}{\textbf{nyl}} & \multicolumn{3}{c}{\textbf{loop1} } & \multicolumn{3}{c}{\textbf{loop2}} \\
\cmidrule(lr){2-4} \cmidrule(lr){5-7} \cmidrule(lr){8-10} \cmidrule(lr){11-13}
& $t_{\text{rel}}$ & $r_{\text{rel}}$ & $t_{\text{abs}}$ & $t_{\text{rel}}$ & $r_{\text{rel}}$ & $t_{\text{abs}}$ & $t_{\text{rel}}$ & $r_{\text{rel}}$ & $t_{\text{abs}}$ & $t_{\text{rel}}$ & $r_{\text{rel}}$ & $t_{\text{abs}}$ \\
& (\%) & (deg/m) & (m) & (\%) & (deg/m) & (m) & (\%) & (deg/m) & (m) & (\%) & (deg/m) & (m) \\
\midrule
\textbf{\revd{Method 1}} & 5.570& 0.129& 1.934& 18.069 & 0.1040  & 103.518  & 14.471 & 0.018 & 288.696 & 15.233 &  0.023 & 401.027 \\
\textbf{\revd{Method 2}} & 1.871& 0.038 & \textbf{0.988}& 2.489& 0.0111& \textbf{6.414}&5.447&0.006& \textbf{104.959} & 4.549& 0.006 & \textbf{52.976 } \\
\textbf{\revd{Method 3}}& 4.13& 0.055& 3.96& 4.62& 0.0184& 17.42& 4.84 & 0.0060& 132.92 & 3.22& 0.0060 & 57.29\\
\textbf{\revd{Method 4}}& 3.56&\textbf{ }0.0369& 2.61& 3.55& 0.0171 & 21.30& 6.09& 0.0082& 227.54&4.09& 0.0097& 59.12\\
\midrule
\textbf{\revd{Method 5}}& 2.79& 0.0511&2.54& 3.10& 0.0120 & 14.34& 4.12 & 0.0065& 68.88 & 3.46& 0.0052& 72.28\\
\textbf{\revd{Method 6}}& 3.02& 0.0448& 2.35& 2.85 & 0.0131& 11.37 & 5.79& 0.0100& 84.88& 4.03& 0.0069& 43.67\\
\bottomrule
\end{tabular}}
\caption{\revdg{Comparison of relative translation (\(t_{\text{rel}}\)), relative rotation (\(r_{\text{rel}}\)), and absolute translation (\(t_{\text{abs}}\)) errors across different datasets and algorithms.  
\textbf{Method 1}: Velocity (ours), \textbf{Method 2}: SLAM (ours), \textbf{Method 3}: 4DRadarSLAM gicp, \textbf{Method 4}: 4DRadarSLAM adpgicp, \textbf{Method 5}: 4DRadarSLAM gicp-lc, \textbf{Method 6}: 4DRadarSLAM adpgicp-lc.  
The results for -lc algorithms are shown for informational purposes only.}}
\label{Table02}
\end{table}

\revd{\subsection{Runtime and memory consumption}}
\revd{The experiments were performed on an HP Victus 16 laptop equipped with 32GB RAM and a 13th generation Intel Core i7-13700H processor.}
\revd{The software implementation was divided into two different processes: the \textbf{Cloud Processing Node} in charge of the point-cloud filtering and ego-velocity computation, and the \textbf{Optimizer Node} dealing with the graph computation, update and optimization. The complete system maintains a moderate memory footprint, with CPU usage consistently reaching 4.2\% for the \textbf{Optimizer Node} and around 0.9\% for the \textbf{Cloud Processing Node}. Memory consumption remains stable at approximately 49.5 MB (0.15 \% of total system memory) for the optimizer and 11.2MB (0.03\% of total system memory) for the cloud processing.}

\revd{Regarding runtime, the system has been tested across several datasets, with execution times generally following similar patterns for both nodes. Thus, the \textbf{Cloud Processing Node} operates at an average execution time of $8.91 \pm 2.27$ miliseconds (see Table~\ref{tab:execution_times}). More specifically, Pre-Processing accounts for $68.3\%$ of the total execution time, Ego-Velocity Estimation takes up $28.36\%$, and Cloud Filtering represents $3.26\%$.
} 

\revd{The \textbf{Optimizer Node} has a more complex execution pattern, as it depends on whether a new keyframe is added (see Table~\ref{tab:execution_times2}). When no keyframe is added, the process involves only Direct Odometry Estimation, which takes an average of $0.12 \pm 0.06$ miliseconds, resulting in minimal computational overhead. However, when a new keyframe is added, the optimization process significantly increases the execution time to an average of $713.52 \pm 334.10$ miliseconds. This is primarily due to the Update Constraints process, which accounts for $99.86\%$ of the total time. The remaining computational cost is distributed among Direct Odometry Estimation ($0.04\%$), Graph Creation ($0.02\%$), and Graph Optimization ($0.07\%$). }

\revd{Although optimization introduces a delay exceeding the radar's operational frequency of $10$ Hz, this impact is mitigated by the fact that keyframe additions occur less frequently than direct odometry updates. As a result, the overall runtime of the optimizer node averages $29.70 \pm 157.00$ miliseconds, ensuring that the delay is balanced across multiple iterations.
}

\begin{table}[t!]
    \centering
    \renewcommand{\arraystretch}{1.3} 
    \setlength{\arrayrulewidth}{0.3mm} 
    \setlength{\tabcolsep}{6pt} 

    \begin{tabular}{l|c c}
         \rowcolor{gray!20} & \textbf{Mean (ms)}& \textbf{Std. Dev. (ms)}\\ 
        \hline
        
        \rowcolor{gray!20} \textbf{Total Execution Process}  & \textbf{8.91}& \textbf{2.27}\\
        
        \rowcolor{white} Pre-Processing & 6.09& 1.62\\
        \rowcolor{white} Ego-Vel Estimation & 6.09& 1.72\\
       \rowcolor{white} Cloud Filtering & 0.29& 0.08\\

    \end{tabular}

    \caption{\revd{Execution time analysis of the \textbf{Cloud Processing Node}}}
    \label{tab:execution_times}
\end{table}

\begin{table}[t!]
    \centering
    \renewcommand{\arraystretch}{1.3} 
    \setlength{\arrayrulewidth}{0.3mm} 
    \setlength{\tabcolsep}{6pt} 

    \begin{tabular}{l|c c}
        \rowcolor{gray!20} & \textbf{Mean (ms)}& \textbf{Std. Dev. (ms)}\\ 
        \hline
        \rowcolor{white} \textbf{Total Execution Process} & \textbf{29.70}& \textbf{157.00}\\ 
        \hline
        \rowcolor{gray!20} \textbf{When New KeyFrame Added}  & 713.52& 334.05\\
        
        \rowcolor{white} Direct Odometry Estimation & 0.30& 0.072\\
        \rowcolor{white} Update Constraints (GICP) & 712.50& 334.05\\
       \rowcolor{white} Create Graph & 0.14& 0.02\\
        \rowcolor{white} Optimize Graph & 0.52& 0.11\\
        
        \hline
        \rowcolor{gray!20} \textbf{When No KeyFrame Added} &  0.12& 0.06\\
    \end{tabular}

    \caption{\revd{Execution time analysis of the \textbf{Optimizer Node}}}
    \label{tab:execution_times2}
\end{table}

\section{Conclusions}

In this paper, a loosely coupled 4D Radar-Inertial Odometry system for terrestrial vehicles is presented, specifically addressing both holonomic and non-holonomic vehicles.

An enhanced ego-velocity estimation system for the vehicle has been proposed, relying on the Doppler effect provided by the radar and taking into account the vehicle's motion model to reduce the impact of radar noise. This system combines radar information with data from the IMU. The resulting velocities have been analyzed, and their impact has been studied through absolute errors, showing that achieving a more accurate velocity estimation leads to significantly lower errors.
This velocity has been incorporated into the proposed odometry system, which is based on graph optimization over a sliding window with a defined number of poses, establishing relationships between them as a mesh to reduce the impact of noise and data inconsistencies, thereby creating a more robust system. The obtained results successfully reduce the drift present in direct odometry, showing improved or comparable performance when tested against other existing methods using the same dataset.

As future work, a tightly coupled approach between the radar and IMU is intended, not only allowing the IMU to influence radar data but also \revd{ enabling radar-derived egovelocity to enhance IMU orientation estimation.} The aim is to improve the IMU preintegration system by allowing the estimation of parameters such as gyroscope and accelerometer bias through additional information provided to the implemented filter, thus enhancing the accuracy of the system’s orientation evolution.

This idea arises from a study conducted on the IMU data from the employed dataset, where non-negligible errors were observed in the pitch and roll angles. Although these angles are generally quite reliable, persistent discrepancies were noted when compared to the ground truth, which, while small, were significant enough to impact subsequent trajectory estimation. With the proposed implementation, the goal is to enhance vehicle orientation information by tightly coupling these two sensors.

Another future direction involves implementing a loop closure system, which comprises two main stages. The first stage is a back-end system responsible for selecting candidate poses for potential loop closures by analyzing both the position and orientation of each pose relative to those stored in the trajectory up to that point. In the second stage, if a valid candidate is identified, the system performs a global optimization over the entire trajectory to incorporate the loop closure constraint.

Loop closure systems for radar data must address challenges such as the noise and sparsity of radar measurements, particularly affecting the vertical component, as discussed in this report. Addressing these challenges is crucial for enhancing the robustness of radar-based odometry, allowing it to better handle long-term drift and improve the overall accuracy of the system.

\section*{Acknowledgements}
Funding for open access publishing: Universidad Pablo de Olavide/CBUA. This work was partially supported by the grants INSERTION (PID2021-127648OB-C31) and NORDIC (TED2021-132476B-I00), both funded by the ``Agencia Estatal de Investigación -- Ministerio de Ciencia, Innovación y Universidades'' and the ``European Union NextGenerationEU/PRTR''.









\begin{thebibliography}{31}

\bibitem{legoloam2018}
T. Shan and B. Englot.
\textit{LeGO-LOAM: Lightweight and Ground-Optimized Lidar Odometry and Mapping on Variable Terrain},
in \textit{IEEE/RSJ International Conference on Intelligent Robots and Systems (IROS)},
2018, pp. 4758--4765.

\bibitem{yan2023gs}
C. Yan, D. Qu, D. Xu, B. Zhao, Z. Wang, D. Wang, and X. Li.
\textit{GS-SLAM: Dense Visual SLAM with 3D Gaussian Splatting},
in \textit{CVPR},
2024.

\bibitem{openvslam2019}
S. Sumikura, M. Shibuya, and K. Sakurada.
\textit{OpenVSLAM: A Versatile Visual SLAM Framework},
\textit{arXiv preprint arXiv:1910.01122},
2019, \url{https://arxiv.org/abs/1910.01122}.

\bibitem{burnett2022}
K. Burnett, Y. Wu, D. J. Yoon, A. P. Schoellig, and T. D. Barfoot.
\textit{Are We Ready for Radar to Replace Lidar in All-Weather Mapping and Localization?},
\textit{IEEE Robotics and Automation Letters},
vol. 7, no. 4, pp. 10328--10335,
2022, Oct.,
doi: \href{https://doi.org/10.1109/LRA.2022.3192885}{10.1109/LRA.2022.3192885}.

\bibitem{zywanowski2020}
K. Żywanowski, A. Banaszczyk, and M. R. Nowicki.
\textit{Comparison of camera-based and 3D LiDAR-based place recognition across weather conditions},
in \textit{2020 16th International Conference on Control, Automation, Robotics and Vision (ICARCV)},
2020, pp. 886--891, Shenzhen, China,
doi: \href{https://doi.org/10.1109/ICARCV50220.2020.9305429}{10.1109/ICARCV50220.2020.9305429}.

\bibitem{ITSC2013}
D. Kellner, M. Barjenbruch, J. Klappstein, J. Dickmann, and K. Dietmayer.
\textit{Instantaneous ego-motion estimation using Doppler radar},
in \textit{16th International IEEE Conference on Intelligent Transportation Systems (ITSC 2013)},
2013, pp. 869--874,
doi: \href{https://doi.org/10.1109/ITSC.2013.6728341}{10.1109/ITSC.2013.6728341}.

\bibitem{9760734}
I. Bilik.
\textit{Comparative Analysis of Radar and Lidar Technologies for Automotive Applications},
\textit{IEEE Intelligent Transportation Systems Magazine},
vol. 15, no. 1, pp. 244--269,
2023,
doi: \href{https://doi.org/10.1109/MITS.2022.3162886}{10.1109/MITS.2022.3162886}.

\bibitem{10530463}
Y. Cheng, M. Jiang, and Y. Liu.
\textit{MS-VRO: A Multistage Visual-Millimeter Wave Radar Fusion Odometry},
\textit{IEEE Transactions on Robotics},
vol. 40, pp. 3004--3023,
2024,
doi: \href{https://doi.org/10.1109/TRO.2024.3400941}{10.1109/TRO.2024.3400941}.

\bibitem{Chan_2023}
P. H. Chan, S. R. Shahbeigi Roudposhti, X. Ye, and V. Donzella.
\textit{A noise analysis of 4D RADAR: robust sensing for automotive?},
\textit{IEEE TechRxiv Preprint},
2023,
doi: \href{https://doi.org/10.36227/techrxiv.24517249.v1}{10.36227/techrxiv.24517249.v1}.

\bibitem{hong2020}
Z. Hong, Y. Petillot, and S. Wang.
\textit{RadarSLAM: Radar based Large-Scale SLAM in All Weathers},
in \textit{2020 IEEE/RSJ International Conference on Intelligent Robots and Systems (IROS)},
2020, pp. 5164--5170,
doi: \href{https://doi.org/10.1109/IROS45743.2020.9341287}{10.1109/IROS45743.2020.9341287}.

\bibitem{adolfsson2021cfearRadarodometryconservative}
D. Adolfsson, M. Magnusson, A. Alhashimi, A. J. Lilienthal, and H. Andreasson.
\textit{CFEAR Radarodometry -- Conservative Filtering for Efficient and Accurate Radar Odometry},
2021, \textit{arXiv preprint},
\url{https://arxiv.org/abs/2105.01457}.

\bibitem{article}
U. Chipengo, P. Krenz, and S. Carpenter.
\textit{From Antenna Design to High Fidelity, Full Physics Automotive Radar Sensor Corner Case Simulation},
\textit{Modelling and Simulation in Engineering},
vol. 2018, pp. 1--19,
2018, Dec.,
doi: \href{https://doi.org/10.1155/2018/4239725}{10.1155/2018/4239725}.

\bibitem{4DRadarSLAM}
J. Zhang, H. Zhuge, Z. Wu, G. Peng, M. Wen, Y. Liu, and D. Wang.
\textit{4DRadarSLAM: A 4D Imaging Radar SLAM System for Large-scale Environments based on Pose Graph Optimization},
in \textit{2023 IEEE International Conference on Robotics and Automation (ICRA)},
2023, pp. 8333--8340,
doi: \href{https://doi.org/10.1109/ICRA48891.2023.10160670}{10.1109/ICRA48891.2023.10160670}.

\bibitem{article2}
B. Wang, Y. Zhuang, and N. El-Bendary.
\textit{4D RADAR/IMU/GNSS Integrated Positioning and Mapping for Large-Scale Environments},
\textit{The International Archives of the Photogrammetry, Remote Sensing and Spatial Information Sciences},
vol. XLVIII-1/W2-2023, pp. 1223--1228,
2023,
doi: \href{https://doi.org/10.5194/isprs-archives-XLVIII-1-W2-2023-1223-2023}{10.5194/isprs-archives-XLVIII-1-W2-2023-1223-2023}.

\bibitem{9235254}
C. Doer and G. Trommer.
\textit{An EKF Based Approach to Radar Inertial Odometry},
in \textit{2020 IEEE International Conference on Multisensor Fusion and Integration for Intelligent Systems (MFI)},
2020, pp. 152--159.

\bibitem{callmer2011}
J. Callmer, D. Törnqvist, F. Gustafsson, H. Svensson, and P. Carlbom.
\textit{Radar SLAM using visual features},
\textit{EURASIP Journal on Advances in Signal Processing},
vol. 2011, no. 71,
2011,
doi: \href{https://doi.org/10.1186/1687-6180-2011-71}{10.1186/1687-6180-2011-71}.

\bibitem{9470842}
C. Doer and G. F. Trommer.
\textit{Yaw aided Radar Inertial Odometry using Manhattan World Assumptions},
in \textit{2021 28th Saint Petersburg International Conference on Integrated Navigation Systems (ICINS)},
2021, pp. 1--9,
doi: \href{https://doi.org/10.23919/ICINS43216.2021.9470842}{10.23919/ICINS43216.2021.9470842}.

\bibitem{shin2024multirobotrelativeposeestimation}
K. Shin, H. Sim, S. Nam, Y. Kim, J. Hu, and K. K. Kim.
\textit{Multi-Robot Relative Pose Estimation in SE(2) with Observability Analysis: A Comparison of Extended Kalman Filtering and Robust Pose Graph Optimization},
2024, \textit{arXiv preprint},
\url{https://arxiv.org/abs/2401.15313}.

\bibitem{mscrad4r}
{MSCRad4r Team}.
\textit{MSCRad4r Dataset},
2023,
\url{https://mscrad4r.github.io/}, accessed 2024-10-30.

\bibitem{safa2022fusingeventbasedcameraradar}
A. Safa, T. Verbelen, I. Ocket, A. Bourdoux, H. Sahli, F. Catthoor, and G. Gielen.
\textit{Fusing Event-based Camera and Radar for SLAM Using Spiking Neural Networks with Continual STDP Learning},
2022, \textit{arXiv preprint},
\url{https://arxiv.org/abs/2210.04236}.

\bibitem{doer2020ekf}
C. Doer and G. F. Trommer, ``An EKF Based Approach to Radar Inertial Odometry,'' in \textit{2020 IEEE International Conference on Multisensor Fusion and Integration for Intelligent Systems (MFI)}, 2020, pp. 152--159.

\bibitem{lim-2023-icra}
H. Lim, K. Han, G. Shin, G. Kim, S. Hong, and H. Myung, ``ORORA: Outlier-robust radar odometry,'' in \textit{Proceedings of the IEEE International Conference on Robotics and Automation (ICRA)}, 2023, pp. 2046--2053.

\bibitem{zhang2018tutorial}
Z. Zhang and D. Scaramuzza, ``A Tutorial on Quantitative Trajectory Evaluation for Visual(-Inertial) Odometry,'' in \textit{IEEE/RSJ International Conference on Intelligent Robots and Systems (IROS)}, 2018.

\bibitem{min2021voldorslamtimesfeaturebaseddirect}
Z. Min and E. Dunn, ``VOLDOR-SLAM: For the Times When Feature-Based or Direct Methods Are Not Good Enough,'' \textit{arXiv preprint}, 2021. Available: \url{https://arxiv.org/abs/2104.06800}.

\bibitem{zhang2023ntu4dradlm4dradarcentricmultimodal}
J. Zhang, H. Zhuge, Y. Liu, G. Peng, Z. Wu, H. Zhang, Q. Lyu, H. Li, C. Zhao, D. Kircali, S. Mharolkar, X. Yang, S. Yi, Y. Wang, and D. Wang, ``NTU4DRadLM: 4D Radar-centric Multi-Modal Dataset for Localization and Mapping,'' \textit{arXiv preprint arXiv:2309.00962}, 2023. [Online]. Available: \url{https://arxiv.org/abs/2309.00962}.


\bibitem{7271006}
D. Holz, A. E. Ichim, F. Tombari, R. B. Rusu, and S. Behnke, ``Registration with the Point Cloud Library: A Modular Framework for Aligning in 3-D,'' \textit{IEEE Robotics \& Automation Magazine}, vol. 22, no. 4, pp. 110--124, 2015. doi: \href{https://doi.org/10.1109/MRA.2015.2432331}{10.1109/MRA.2015.2432331}.

\bibitem{wu2024efear4degovelocityfilteringefficient}
X. Wu, Y. Chen, Z. Li, Z. Hong, and L. Hu, ``EFEAR-4D: Ego-Velocity Filtering for Efficient and Accurate 4D radar Odometry,'' \textit{arXiv preprint}, 2024. Available: \url{https://arxiv.org/abs/2405.09780}.

\bibitem{6728341}
D. Kellner, M. Barjenbruch, J. Klappstein, J. Dickmann, and K. Dietmayer, ``Instantaneous ego-motion estimation using Doppler radar,'' in \textit{16th International IEEE Conference on Intelligent Transportation Systems (ITSC 2013)}, 2013, pp. 869--874. doi: \href{https://doi.org/10.1109/ITSC.2013.6728341}{10.1109/ITSC.2013.6728341}.

\bibitem{boekema2024vodp}
H. J.-H. Boekema, B. K. W. Martens, J. F. P. Kooij, and D. M. Gavrila, ``Multi-class Trajectory Prediction in Urban Traffic using the View-of-Delft Prediction Dataset,'' \textit{IEEE Robotics and Automation Letters}, vol. 9, no. 5, pp. 4806--4813, 2024. doi: \href{https://doi.org/10.1109/LRA.2024.3385693}{10.1109/LRA.2024.3385693}.

\bibitem{lu2020milliegosinglechipmmwaveradar}
C. X. Lu, M. R. U. Saputra, P. Zhao, Y. Almalioglu, P. P. B. de Gusmao, C. Chen, K. Sun, N. Trigoni, and A. Markham, ``milliEgo: Single-chip mmWave Radar Aided Egomotion Estimation via Deep Sensor Fusion,'' \textit{arXiv preprint}, 2020. Available: \url{https://arxiv.org/abs/2006.02266}.

\bibitem{10100861}
Y. Zhuang, B. Wang, J. Huai, and M. Li, ``4D iRIOM: 4D Imaging Radar Inertial Odometry and Mapping,'' \textit{IEEE Robotics and Automation Letters}, vol. 8, no. 6, pp. 3246--3253, 2023. doi: \href{https://doi.org/10.1109/LRA.2023.3266669}{10.1109/LRA.2023.3266669}.

\bibitem{article}
U. Chipengo, P. Krenz, and S. Carpenter, ``From Antenna Design to High Fidelity, Full Physics Automotive Radar Sensor Corner Case Simulation,'' \textit{Modelling and Simulation in Engineering}, vol. 2018, pp. 1--19, 2018. doi: \href{https://doi.org/10.1155/2018/4239725}{10.1155/2018/4239725}.
\bibitem{ceres-solver}
S. Agarwal, K. Mierle, and Others,  
\textit{Ceres Solver},  
Available: \url{http://ceres-solver.org},  
Accessed: April 1, 2024.

\end{thebibliography}
\end{document}